\documentclass[10pt,twocolumn,letterpaper]{article}

\usepackage[pagenumbers]{cvpr} 
\usepackage{comment}
\usepackage{soul}
\usepackage{multirow}
\usepackage{multicol}
\usepackage{float}
\usepackage{cuted}
\usepackage{lipsum}
\usepackage{marvosym}

\definecolor{tx_color}{RGB}{200,42,40}
\definecolor{delete_red}{RGB}{255,72,32}
\definecolor{fweS_color}{RGB}{255,0,0}
\definecolor{weBc_color}{RGB}{0,255,0}
\definecolor{jaza_color}{RGB}{0,0,255}

\newcommand{\OURS}{DashGaussian\xspace}%
\newcommand{\CPLX}{optimization complexity}
\newcommand{\ACC}{45.7\%}

\usepackage{amsmath}
\DeclareMathOperator*{\argmin}{argmin}

\definecolor{cvprblue}{rgb}{0.21,0.49,0.74}
\usepackage[pagebackref,breaklinks,colorlinks,allcolors=cvprblue]{hyperref}
\usepackage{xcolor}

\def\paperID{4144} 
\def\confName{CVPR}
\def\confYear{2025}

\title{\OURS{}: Optimizing 3D Gaussian Splatting in 200 Seconds}

\author{
Youyu Chen\textsuperscript{1\#}~~~~~Junjun Jiang\textsuperscript{1\Cross}~~~~~Kui Jiang\textsuperscript{1}~~~~~Xiao Tang\textsuperscript{2}\\
Zhihao Li\textsuperscript{2}~~~~~Xianming Liu\textsuperscript{1}~~~~~Yinyu Nie\textsuperscript{2}
\\\vspace{-3pt} \normalsize \textsuperscript{1}Harbin Institute of Technology 
\quad \textsuperscript{2}Huawei Noah's Ark Lab
\vspace{-10pt}
}

\begin{document}



\begin{figure}
    \vspace{-4mm}
    \twocolumn[{
        \renewcommand\twocolumn[1][]{#1}
        \maketitle
        
        \centering
        \setlength\tabcolsep{0pt}
        \includegraphics[width=\linewidth]{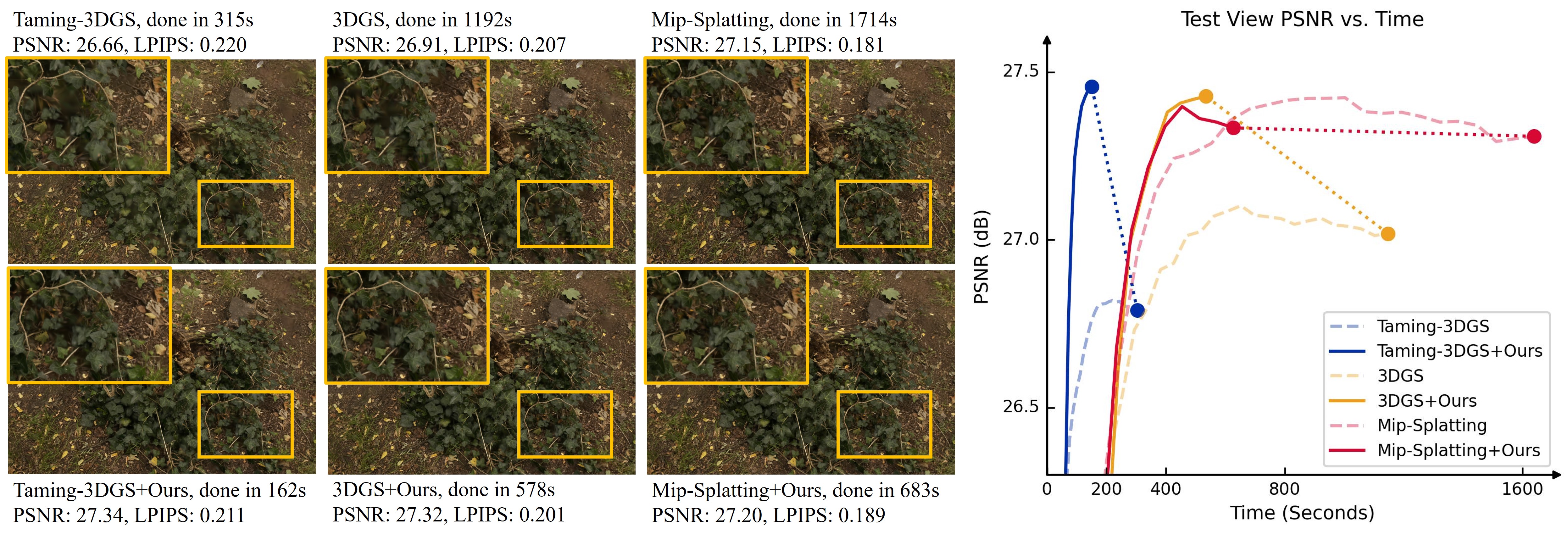}
        \vspace{-4mm}
        \caption{
            We propose \OURS{}, a fast 3D Gaussian Splatting (3DGS) optimization method that can be easily plugged into existing 3DGS backbones.
            \OURS{} significantly boosts the training speed of various 3DGS backbones by \textbf{\ACC{}} on average without trading off rendering quality. 
            Equipping \OURS{} to prior-art 3DGS methods, we reduce the optimization time of a 3DGS model with millions of primitives to \textbf{200 seconds} on a consumer-grade GPU.
            The figures above show the scene ``stump'' in the Mip-NeRF 360 dataset.
        }
        \vspace{4mm}
        \label{fig:teaser}
    }]
\end{figure}

\begin{abstract}
3D Gaussian Splatting (3DGS) renders pixels by rasterizing Gaussian primitives, where the rendering resolution and the primitive number, concluded as the \CPLX{}, dominate the time cost in primitive optimization. 
In this paper, we propose \OURS{}, a scheduling scheme over the \CPLX{} of 3DGS that strips redundant complexity to accelerate 3DGS optimization. 
Specifically, we formulate 3DGS optimization as progressively fitting 3DGS to higher levels of frequency components in the training views, and propose a dynamic rendering resolution scheme that largely reduces the optimization complexity based on this formulation. 
Besides, we argue that a specific rendering resolution should cooperate with a proper primitive number for a better balance between computing redundancy and fitting quality, where we schedule the growth of the primitives to synchronize with the rendering resolution. 
Extensive experiments show that our method accelerates the optimization of various 3DGS backbones by \ACC{} on average while preserving the rendering quality. 
Project page is available at \url{dashgaussian.github.io}.
\vspace{0mm}
\end{abstract}
\newcommand\blfootnote[1]{%
  \begingroup
  \renewcommand\thefootnote{}\footnote{#1}%
  \addtocounter{footnote}{-1}%
  \endgroup
}
\blfootnote{$^\#$This work was done during he was an intern at Huawei.}
\blfootnote{$^\text{\Cross}$Corresponding author. E-mail: \tt{jiangjunjun@hit.edu.cn}.}
\vspace{-11mm}
\section{Introduction}
\label{sec:intro}

Novel view synthesis has long been investigated as an important branch of 3D scene reconstruction. Neural Radiance Field (NeRF)~\cite{10.1145/3503250} as a milestone work has largely improved the visual quality of novel views, but at the cost of days to optimize a single scene. 
Though many efforts~\cite{mueller2022instant, liu2020neural, Fridovich-Keil_2022_CVPR, Chen_2023_CVPR, 10.1007/978-3-031-19824-3_20, 10.1145/3658193, adaptiveshells2023, chen2024mesh2nerfdirectmeshsupervision} have been made to improve the efficiency, NeRF-based methods still face challenges in balancing the optimization time and rendering quality. 
In contrast, 3D Gaussian Splatting (3DGS)~\cite{kerbl3Dgaussians} utilizes Gaussian primitives to model a scene, achieving equivalent novel view rendering quality as NeRF while reducing the optimization time of one scene to tens of minutes. 
Nevertheless, there remains a high demand for further acceleration to enable 3DGS applications on devices with limited computational power or in large-scale, time-intensive scene reconstruction tasks~\cite{Lin_2024_CVPR, liu2024citygaussian, yuchen2024dogaussian, Li_2023_ICCV}.

Given a specific 3DGS backbone, there are multiple strategies to accelerate its optimization, which can be collected into two major categories. 
From an engineering perspective, some works~\cite{durvasula2023distwarfastdifferentiablerendering, feng2024flashgsefficient3dgaussian, mallick2024taming3dgshighqualityradiance, ye2024gsplatopensourcelibrarygaussian, deng2025efficientdensitycontrol3d} accelerate the optimization by refining the rendering pipeline to enable high-efficiency computation. 
From an algorithmic perspective, some works~\cite{10.1145/3651282,fang2024minisplattingrepresentingscenesconstrained} propose to prune redundant Gaussian primitives after primitive densification.
While these algorithmic methods perform effectively, the rendering quality noticeably declines when a considerable number of parameters are excluded to speed up the optimization.

\begin{figure*}[t]
    \centering
    \captionsetup{singlelinecheck=false}
    \includegraphics[width=\linewidth]{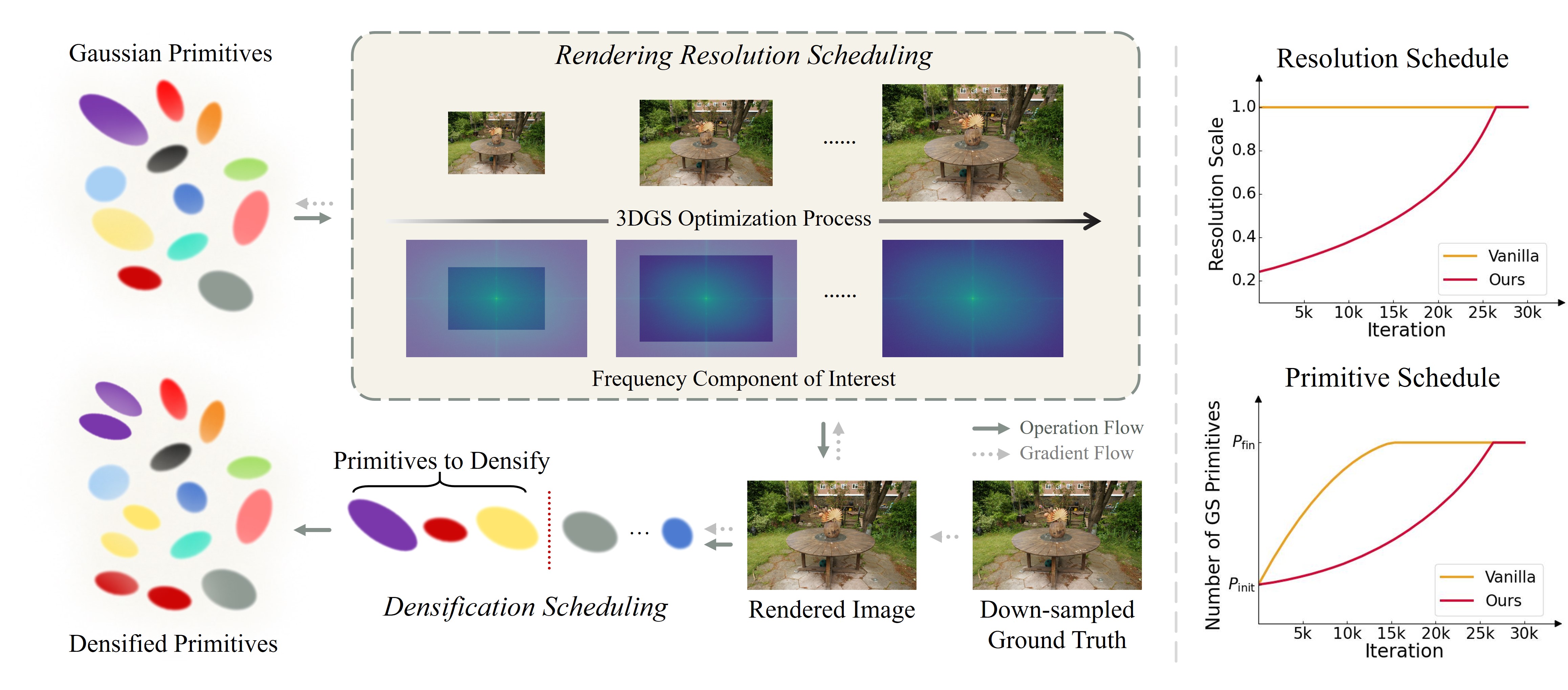}
    \caption{
        \textbf{The pipeline of \OURS{}.} 
        \OURS{} determines the rendering resolution for each 3DGS optimization step with our resolution scheduling method. 
        The insight of the resolution scheduling is to gradually fit 3DGS to higher level of frequency components in the training views throughout the entire optimization process. 
        By directing the downsampling of training views with our scheduler, we significantly reduce the time cost for 3DGS optimization while preserving the rendering quality. 
        We further manage the growth of Gaussian primitives, which cooperates with the scheduled rendering resolution. It prevents possible over-densification issues during the low-resolution optimization phase and further accelerates the optimization with suppressed primitive growth. 
    }
    \label{fig:framework}
\end{figure*}
Instead of eliminating redundant Gaussian primitives, we pursue an effective strategy for allocating computational resources, which smartly leverages each Gaussian primitive to be more time-efficiently optimized.
This is based on three key observations. 
First, in 3DGS optimization, computational resources are primarily consumed by three main operations — forward rendering, gradient back-propagation, and primitive update — all of which have their computational complexity commonly relying on the rendering resolution and the number of Gaussian primitives. 
Second, rendering high-resolution images at the beginning of optimization is thriftless because the primitives are sparse at this stage. 
Third, the late stage of the optimization consumes a huge portion of time cost because of the large primitive number while bringing limited rendering quality improvement (see the curves of backbones in \cref{fig:teaser}).
These observations inspire us to propose a more effective allocation of computational resources for current 3DGS methods.
For simplicity, we refer to the rendering resolution and the number of primitives collectively as the `\CPLX{}'.
By manipulating the \CPLX{} on different optimization stages, one can arbitrarily customize the computational resource distribution. 
Naively reducing \CPLX{} can promisingly achieve acceleration though, it usually trades off the rendering quality.

One way to manipulate the \CPLX{} is to reduce the rendering resolution in the early stage of 3DGS optimization. 
However, simply reducing the resolution in training results in 3D aliasing~\cite{Yu2024MipSplatting, 10.1007/978-3-031-72643-9_17}, which sacrifices the rendering fidelity in the original resolution. 
Nevertheless, according to image processing theory, when comparing two images — one downsampled from the other — the key difference between them lies in the residual frequency components involved in the high-resolution image but absent in the low-resolution one~\cite{10.5555/1076432}.
This indicates that increasing the rendering resolution during the optimization process is equivalent to gradually fitting 3DGS to higher levels of frequency components in the training views (see \cref{fig:framework}). 
Inspired by this, we propose a scene-adaptive frequency-based resolution scheduling scheme that progressively increases the rendering resolution throughout the entire optimization process while ensuring its fidelity at the end. 

Given the scheduled rendering resolution, we point out that a specific rendering resolution should be paired with an appropriate number of Gaussian primitives to better balance the computational redundancy and fitting quality of 3DGS. 
To this end, we design a primitive number scheduler that tames the primitives to grow simultaneously with the rendering resolution by coupling them with a numerical constraint. 
This scheduler produces a concave-up curve depicting the growth of the primitive, which spans over the entire optimization process (see \cref{fig:framework}). 
Such a concave-up growth curve possesses significant advantage over the ones in existing works that are either concave-down~\cite{kerbl3Dgaussians, mallick2024taming3dgshighqualityradiance} or reach their peak at half of the optimization iterations~\cite{kerbl3Dgaussians, bulò2024revisingdensificationgaussiansplatting}. 
Moreover, we introduce a momentum-based primitive budgeting method to adaptively estimate the maximum number of primitives in the growth curve, rather than relying on dataset priors as in existing methods~\cite{fang2024minisplattingrepresentingscenesconstrained, mallick2024taming3dgshighqualityradiance, bulò2024revisingdensificationgaussiansplatting, NEURIPS2024_93be245f}.

We integrate the proposed resolution scheduler and the primitive scheduler, presenting \OURS{}, which significantly accelerates the optimization process across various 3DGS backbones by \ACC{} on average without trading off the rendering quality. To summarize our contributions, 

\begin{itemize}
    \item 
    We speed up the optimization of 3DGS by reasonably distributing the computational resources throughout the optimization process. 
    \item 
    We propose a scheduling method to corporately control the growth of rendering resolution and Gaussian primitives, ensuring that the rendering quality is maintained.
    \item 
    We design an adaptive Gaussian primitive budgeting method to determine the primitive number demanded by the scene, releasing primitive growth scheduling from appointing the final primitive number heuristically. 
\end{itemize}

\vspace{2mm}
\section{Related Work}
\label{sec:related}

\paragraph{Novel View Synthesis.}
Among the vast amount of investigations on novel view synthesis~\cite{seitz1999photorealistic, 10.1145/1401132.1401175, remondino2006image, buehler2001unstructured, fitzgibbon2005image}, Neural Radiance Fields (NeRF)~\cite{10.1145/3503250} have achieved particular success, but at the cost of tens of hours to optimize for a single scene~\cite{10.1007/978-3-030-58542-6_42, Barron_2021_ICCV, DeepBlending2018, 10.1145/3503250, mueller2022instant, https://doi.org/10.1111/cgf.14507}. 
Although there have been many works focusing on improving NeRF's efficiency~\cite{mueller2022instant, liu2020neural, Fridovich-Keil_2022_CVPR, Chen_2023_CVPR, 10.1007/978-3-031-19824-3_20, adaptiveshells2023, zhong2024cvtxrfcontrastiveinvoxeltransformer}, they still struggle with balancing the optimization efficiency and rendering quality. 
Alternatively, 3D Gaussian Splatting (3DGS)~\cite{kerbl3Dgaussians} emerges as a solution that possesses comparable rendering quality to NeRF together with significant advancement in terms of the speed of both optimization and rendering. 
The efficiency of 3DGS benefits from its GPU-friendly rendering pipeline, which largely reduces the computational burden for its optimization. 
3DGS has been widely applied in different tasks, such as large scene reconstruction~\cite{Lin_2024_CVPR, liu2024citygaussian, yuchen2024dogaussian}, digital human~\cite{10.1007/978-3-031-72684-2_8, 10.1145/3664647.3681675}, and 3D segmentation~\cite{10.1007/978-3-031-72670-5_26, hu2024sagdboundaryenhancedsegment3d}. 
Improving the performance of 3DGS is fundamental and significant to benefit the downstream tasks. 

\paragraph{Acceleration for 3DGS Optimization.}
\vspace{-4mm}
Since the success of 3DGS is largely driven by its computational efficiency, it is intuitive to further improve its speed with high-efficiency computation. 
By refining the implementation of forward-pass, backward-pass and parameter update of 3DGS, recent works~\cite{ye2024gsplatopensourcelibrarygaussian, feng2024flashgsefficient3dgaussian, durvasula2023distwarfastdifferentiablerendering, mallick2024taming3dgshighqualityradiance} accelerate the optimization of 3DGS by a large margin. 
Meanwhile, reducing the parameters of 3DGS is also a promising option to accelerate the optimization. 
For example, Fang and Wang~\cite{fang2024minisplattingrepresentingscenesconstrained} propose to reduce the number of Gaussian primitives by pruning the crowded primitives.
Papantonakis \etal~\cite{10.1145/3651282} further propose to strip the redundant appearance descriptors of each primitive to reduce the optimized parameters. 
However, these methods inevitably hinder the rendering quality because of their hard cropping operation to shrink the size of the original 3DGS model. 
In contrast, \OURS{} ingeniously suppresses the optimization complexity and densifies Gaussian primitives without hard cropping, thus holding the potential to preserve the rendering quality. 

\paragraph{Scheduled 3DGS Optimization}
\vspace{-4mm}
3DGS~\cite{kerbl3Dgaussians} densifies the Gaussian primitives, which results in an unpredictable size of the optimized Gaussians. 
To make the densification controllable, Mallick \etal~\cite{mallick2024taming3dgshighqualityradiance} propose to mimic the primitive growth of 3DGS~\cite{kerbl3Dgaussians} with a constructive densification process. 
Bulò \etal~\cite{bulò2024revisingdensificationgaussiansplatting} propose to densify the Gaussian primitive by a constant ratio until the predetermined primitive budget is met. 
On the other hand, 3DGS with Level of Details (LoD)~\cite{shi2024lapisgslayeredprogressive3d, seo2024flodintegratingflexiblelevel} is optimized with multiple rendering resolutions for seamlessly rendering at varying levels of details.
However, these methods are time-consuming in optimization while their levels of resolution are fixed. 
\OURS{} corporately schedules both rendering resolution and primitive growth in an adaptive manner, which significantly boosts the optimization of 3DGS without predetermining the final primitive number heuristically. 

\vspace{2mm}
\section{Preliminary}
\label{sec:prelim}

Given a set of posed images capturing a scene and the corresponding point cloud derived from the structure-from-motion algorithm~\cite{Schonberger_2016_CVPR}, 3DGS optimizes a set of 3D Gaussian primitives $\{\mathcal{G}_i(\mathbf{x}) = e^{-\frac{1}{2}(\mathbf{x}-\mathbf{p}_i)\Sigma_i^{-1}(\mathbf{x}-\mathbf{p}_i)} \}_{i=1}^P$ to represent the scene for novel view synthesis. 
Each 3D Gaussian primitive $\mathcal{G}_i$ is characterized with its center position $\mathbf{p}_i\in \mathbb{R}^3$, covariance matrix defined in the world space $\Sigma_i\in\mathbb{R}^{3\times 3}$, opacity $o_i \in (0, 1)$ and color coefficients $\mathbf{c}_i$. Here, $P$ is the total number of Gaussian primitives used to represent the scene.
To render images of camera views, 3DGS splats these Gaussian primitives onto pixels. The color $\mathbf{C}$ of a pixel $\mathbf{u}$ is calculated as
\begin{equation}
    \label{eq:1}
    \mathbf{C(\mathbf{u})} = \sum_{i=1}^{P} {T_i(\mathbf{u}) \cdot w_i(\mathbf{u})\cdot\mathbf{c}_i}, 
    \thickspace T_i(\mathbf{u}) = \prod_{j=1}^{i-1}(1-w_j(\mathbf{u})), 
\end{equation}
where $w_i(\mathbf{u})=o_i\mathcal{G}_i^\mathrm{2D}(\mathbf{u})$, and $\mathcal{G}_i^\mathrm{2D}$ denotes the 2D Gaussian splatted to the image plane from $\mathcal{G}_i$. 
As the whole process is fully differentiable, $\mathcal{G}_i$ can be optimized by back-propagating the gradients from losses between renderings and training images. 
The number of rendered pixels, decided by the rendering resolution, indicates the times \cref{eq:1} is processed, thus directly affecting the optimization time.
The original 3DGS algorithm~\cite{kerbl3Dgaussians}~further incorporates a densification operation during the optimization process, which identifies those ${\mathcal{G}_i}$ in under-reconstructed regions and then clones or splits them to better represent the scene.
As the Gaussian primitive number $P$ increases, the computational burden in \cref{eq:1} also grows, which directly impacts optimization efficiency.
We collectively refer to the rendering resolution and the number of primitives as \CPLX{}, and our goal is to reduce the overall \CPLX{} to accelerate 3DGS optimization.
\vspace{2mm}
\section{DashGaussian}
\label{sec:method}

In this section, we introduce how \OURS{} strips the optimization complexity to accelerate 3DGS optimization without trading off rendering quality. 
\cref{sec:method:formulate} reviews the aliasing issue in image downsampling, and proposes to formulate optimizing 3DGS with increasing rendering resolution as fitting to different level of frequency components. 
In \cref{sec:method:resosche}, we introduce our frequency-guided resolution scheduling strategy, which adaptively builds the resolution schedule for each scene based on the frequency component of training views. 
In \cref{sec:method:primsche}, we introduce our resolution-guided primitive scheduling scheme, which tames the number of Gaussian primitives to grow simultaneously with the rendering resolution, establishing a reasonable cooperation between the rendering resolution and the primitive number. 

\vspace{1mm}
\subsection{Problem Formulation}
\label{sec:method:formulate}

Optimizing 3DGS with images in different resolutions while maintaining high rendering fidelity is non-trivial. 
More concretely, we discuss the case where 3DGS is supervised with downsampled low-resolution (LR) images while aiming to render the original high-resolution (HR) ones. 
We investigate this problem as it inevitably leads to 3D aliasing~\cite{Yu2024MipSplatting}, making it formidable to accelerate the optimization with LR supervision while keeping a high fidelity in HR rendering. 
However, viewing 3DGS optimization through the lens of 2D image reconstruction, the frequency theory in image processing offers a fresh perspective.

We first review image downsampling, which is applied to the training views to decrease their resolution.
Given an image $\mathbf{I} \in \mathbb{R}^{H\times W}$ being downsampled to $\mathbf{I}_r \in \mathbb{R}^{\frac{H}{r} \times \frac{W}{r}}$, we denote their frequency maps with Discrete Fourier Transform (DFT) by $\mathbf{F}=\mathrm{DFT}(\mathbf{I}) \in \mathbb{R}^{H\times W}$ and $\mathbf{F}_r=\mathrm{DFT}(\mathbf{I}_r) \in \mathbb{R}^{\frac{H}{r}\times \frac{W}{r}}$, respectively. 
According to the Nyquist Sampling Theorem, one can never recover $\mathbf{I}$ from $\mathbf{I}_r$ if the sampling frequency of $\mathbf{I}_r$ is less than two times of the maximum frequency contained in $\mathbf{F}$.
Meanwhile, a simple spatial downsampling could result in aliasing issues that disturb the recoverable low-frequency components and introduce artifacts in $\mathbf{I}_r$.
To address this issue, anti-aliasing downsampling methods apply a low-pass filter on $\mathbf{F}$ in different ways before downsampling $\mathbf{I}$. 
The most straightforward way is to center crop $\mathbf{F}$ to the size of $(H/r, W/r)$, scale its numerical value by $1/r^2$, then inversely transform the cropped $\mathbf{F}$ into $\mathbf{I}_r$. 

With the above analysis, we see the key difference between $\mathbf{I}$ and $\mathbf{I}_r$ is the residual high-frequency component that $\mathbf{F}$ embeds while $\mathbf{F}_r$ does not. 
Based on this observation, we reformulate optimizing 3DGS with increasing image resolutions by gradually fitting 3DGS to higher levels of frequency components throughout the entire optimization process (see \cref{fig:framework}).  
In \cref{sec:method:resosche}, we demonstrate that this formulation indicates the extent to which downsampled images can be used to optimize a 3DGS model.

\vspace{1mm}
\subsection{Frequency Guided Resolution Scheduler}
\label{sec:method:resosche}

Based on the discussion in \cref{sec:method:formulate}, we propose to schedule the rendering resolution of 3DGS throughout the optimization process, which adaptively distributes optimization steps to different resolutions based on the amount of information contained in the frequency domain of each resolution. 
Since our scheduling covers the entire optimization process, we can largely reduce the optimization complexity and reduce the overall time cost. 
For simplicity, only two different resolutions $(H, W)$ and $(H/r, W/r)$, whose related notations are defined the same as \cref{sec:method:formulate}, are introduced in the following discussion to optimize 3DGS, where $r\geq 1$.
The LR rendering $(H/r, W/r)$ is applied in the early stage of optimization, which is switched to HR rendering $(H, W)$ by the end. 
Here we suppose the LR training views for supervision are aliasing freely downsampled from the HR training views.

The key problem is when to increase the rendering resolution from low to high, solving which is equivalent to answering how to distribute the total $S$ optimization steps between $\mathbf{F}$ and $\mathbf{F}_r$.
Since $\mathbf{F}_r$ is contained within $\mathbf{F}$ because $\mathbf{I}$ is aliasing freely downsampled to $\mathbf{I}_r$, the frequency component in LR images is always accessible in all the $S$ iterations for supervision. 
The question is to know the `significance' of the HR images and decide the portion they take in the $S$ steps to preserve HR rendering fidelity of 3DGS. 

A reasonable definition of the significance should handle both 2D and 3D aliasing issue during optimization. 
Because 2D aliasing happens in the frequency domain, it indicates a significance definition based on a numerical feature of the frequency map, e.g., the frequency energy which dominates the optimization steps with the LR rendering. 
On the other hand, 3D aliasing has been thoroughly discussed by Mip-Splatting~\cite{Yu2024MipSplatting}. 
Though Mip-Splatting alleviated the 3D aliasing issue, it still encourages optimization in HR for high-fidelity HR rendering. 

To balance these two issues, we define the significance of images under a specific resolution $(h, w)$ as
\begin{equation}
    \label{eq:2}
    \mathcal{X}(\mathbf{F})=\frac{1}{N}\sum_{n=1}^N\sum_{i=1}^h\sum_{j=1}^w{||\mathbf{F}^n(i,j)||_2}, 
\end{equation}
where $\mathbf{F}^n=\mathrm{DFT}(\mathbf{I}^n)$; $\{\mathbf{I}^n\}_{n=1}^N$ are the training views sampled with the resolution of $(h, w)$, and $N$ is the number of training views.
Derived from numerical feature in the frequency domain, this definition particular emphasizes the high-frequency component. 
Intuitively, if $\mathcal{X}(\mathbf{F}) - \mathcal{X}(\mathbf{F}_r)$ is far bigger than $\mathcal{X}(\mathbf{F}_r)$, the optimization steps should be dominated with the HR rendering, and vice versa. 
Following this intuition, we define a fraction function by
\begin{equation}
    \label{eq:3}
    f(\mathbf{F}, \mathbf{F}_r) = \frac{\mathcal{X}(\mathbf{F}) - \mathcal{X}(\mathbf{F}_r)}{\mathcal{X}(\mathbf{F})}, 
\end{equation}
with which we distribute $S\cdot f(\mathbf{F}, \mathbf{F}_r)$ iterations to the HR rendering. 
In other words, we should increase the rendering resolution from $(H/r, W/r)$ to $(H, W)$ at the $s_{r}$-th iteration, where $s_r = S\cdot\left(1-f(\mathbf{F}, \mathbf{F}_r)\right) = S\cdot\mathcal{X}(\mathbf{F}_r)/\mathcal{X}(\mathbf{F})$. 

In practice, we can generalize the above discussion and adapt \cref{eq:3} to the case with multiple intermediate resolutions. 
Given a group of downsampling factors $r_1, r_2, ..., r_m$ that monotonically increase, indicating downsampled images $\mathbf{I}_{r_m}, ..., \mathbf{I}_{r_1}, \mathbf{I}$ are progressively involved for supervision throughout the optimization process, we increase the rendering resolution from $(H/r_i, W/r_i)$ to $(H/r_{i-1}, W/r_{i-1})$ at the $s_{r_i}$-th iteration. 

We correspond the maximal downsampling factor $r_m$ with the one whose $\mathcal{X}(\mathbf{F}_{r_m}) = \frac{1}{a}\mathcal{X}(\mathbf{F})$, where $a$ is a hyper-parameter (defaulted by 4). We linearly sample the rest $\mathcal{X}(\mathbf{F}_{r_1}), ..., \mathcal{X}(\mathbf{F}_{r_{m-1}})$ between $\mathcal{X}(\mathbf{F})$ and $\mathcal{X}(\mathbf{F}_{r_m})$, and solve the corresponding resolution factors $r_1, ..., r_{m-1}$. 
The rendering resolution $r^{(k)}$ for iteration $k$ between $s_{r_i}$ and ${s_{r_{i-1}}}$ is interpolated with the inverse square of $r_i$ and $r_{i-1}$. 
Please refer to the supplementary for more details.

By programming the rendering resolution throughout the optimization process of 3DGS with the scheduling scheme above (see the curve in \cref{fig:framework}), we significantly reduce the optimization time while maintaining or even improving the quality of the optimized 3DGS, which is revealed by the experiments in \cref{sec:exp:ablation}.

\vspace{1mm}
\subsection{Resolution Guided Primitive Scheduler}
\label{sec:method:primsche}

Recent works have revealed the fact that over-dense Gaussian primitives only marginally improves rendering quality, but largely slows down the optimization~\cite{fang2024minisplattingrepresentingscenesconstrained, 10.1145/3651282}. 
To further improve the efficiency of \OURS{}, it is vital for the growth of primitives to avoid over densification in the LR optimization stage and determine a proper number of Gaussian primitives in the optimized 3DGS model.

\begin{table*}[t!]
    \centering
    \footnotesize
    \setlength\tabcolsep{2pt}
    \captionsetup{singlelinecheck=false}
    \caption{
        Quantitative comparisons with existing 3DGS fast optimization methods. 
        With DashGaussian, we finish 3DGS optimization within 200 seconds while achieving significantly higher rendering quality compared with the other methods. 
        ``Ours'' in the table reports the performance of equipping DashGaussian to Taming-3DGS$^1$~\cite{mallick2024taming3dgshighqualityradiance}.
        Time is reported in minutes. 
    }
    \begin{tabular} {l | rrrrr | rrrrr | rrrrr}
        \toprule
        \multirow{2}{*}{Method} 
        & \multicolumn{5}{c}{MipNeRF-360~\cite{Barron_2022_CVPR}} & \multicolumn{5}{c}{Deep Blending~\cite{DeepBlending2018}} & \multicolumn{5}{c}{Tanks and Temples~\cite{Knapitsch2017}} \\
        \cmidrule(l{2pt}r{2pt}){2-6} \cmidrule(l{2pt}r{2pt}){7-11} \cmidrule(l{2pt}r{2pt}){12-16}
        & $\mathrm{N_{GS}}\downarrow$ & PSNR$\uparrow$ & SSIM$\uparrow$ & LPIPS$\downarrow$ & Time$\downarrow$ 
        & $\mathrm{N_{GS}}\downarrow$ & PSNR$\uparrow$ & SSIM$\uparrow$ & LPIPS$\downarrow$ & Time$\downarrow$ 
        & $\mathrm{N_{GS}}\downarrow$ & PSNR$\uparrow$ & SSIM$\uparrow$ & LPIPS$\downarrow$ & Time$\downarrow$ \\
        \midrule
        3DGS~\cite{kerbl3Dgaussians}
                            &          3.12M &          27.72 &          0.826 & \textbf{0.201} &          18.31 
                            &          2.82M &          29.50 &          0.904 & \textbf{0.244} &          17.27 
                            &          1.83M &          23.62 &          0.847 &          0.177 &          10.59 \\
        Reduced-3DGS~\cite{10.1145/3651282}
                            &          1.44M &          27.28 &          0.813 &          0.222 &          15.52 
                            &          0.98M &          29.78 &          0.907 &          0.245 &          13.74 
                            &          0.65M &          23.59 &          0.844 &          0.185 &           8.31 \\
        Mini-Splatting~\cite{fang2024minisplattingrepresentingscenesconstrained}
                            &  \textbf{0.49M} &          27.22 &          0.821 &          0.217 &          14.72 
                            &  \textbf{0.34M} &          29.95 &          0.907 &          0.253 &          12.91 
                            &  \textbf{0.20M} &          23.20 &          0.835 &          0.201 &           8.66 \\
        Taming-3DGS~\cite{mallick2024taming3dgshighqualityradiance}
                            &          2.28M &          27.61 &          0.821 &          0.210 &           5.51 
                            &          2.32M &          29.69 &          0.906 &          0.245 &           4.52 
                            &          1.51M &          23.62 &          0.849 & \textbf{0.174} &           4.02 \\
        \midrule
        DashGaussian (Ours) 
                            &          2.04M & \textbf{27.92} & \textbf{0.826} &          0.208 &  \textbf{3.23} 
                            &          1.94M & \textbf{30.02} & \textbf{0.907} &          0.248 &  \textbf{2.20} 
                            &          1.19M & \textbf{23.97} & \textbf{0.851} &          0.180 &  \textbf{2.62} \\
        \bottomrule
    \end{tabular}
    \label{tab:main-result}
\end{table*}
        

\begin{figure*}[t]
    \centering
    \captionsetup{singlelinecheck=false}
    \includegraphics[width=\linewidth]{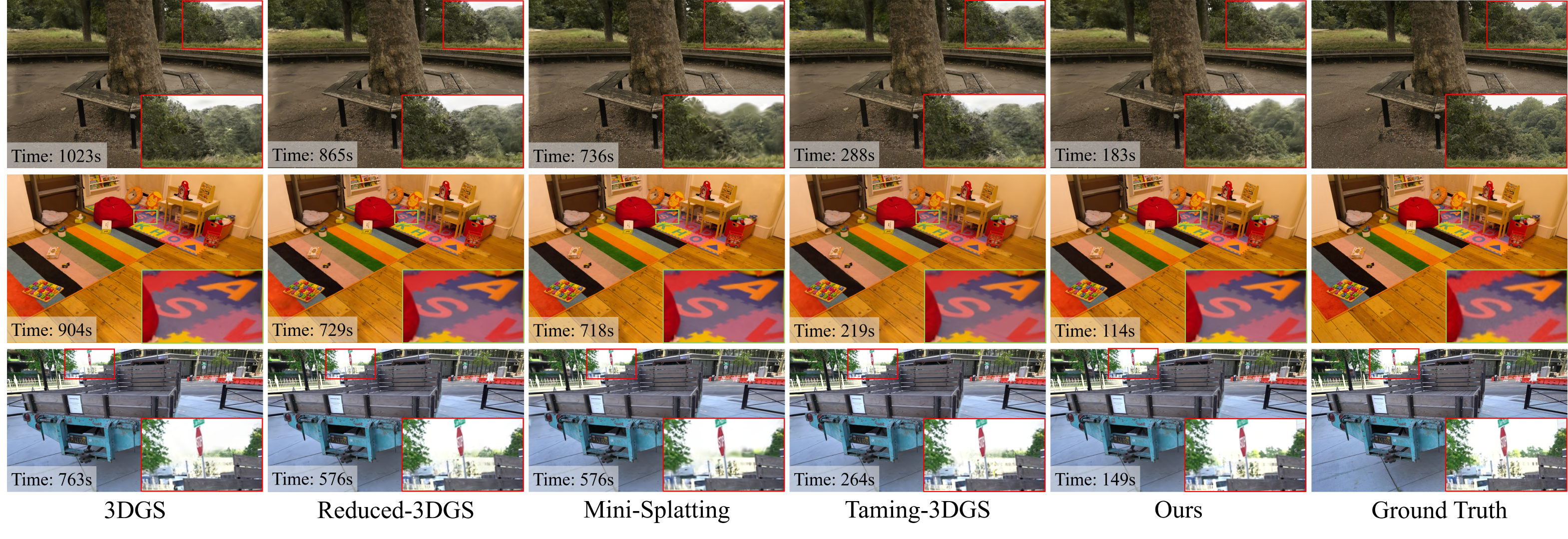}
    \vspace{-6mm}
    \caption{
        Qualitative results of \cref{tab:main-result}.  
        We show the rendering results of the ``treehill'' scene in the Mip-NeRF 360 dataset~\cite{Barron_2022_CVPR}, the ``playroom'' scene in the Deep Blending dataset~\cite{DeepBlending2018}, and the ``truck''scene in the Tanks\&Temples dataset~\cite{Knapitsch2017}, respectively. 
    }
    \label{fig:expmainfigure}
\end{figure*}

\paragraph{Resolution Guided Densification. }
\vspace{-2mm}
Since 3DGS is supervised with images capturing the surface of the scene, which the Gaussian primitives majorly describe~\cite{Yu2024GOF, 10.1145/3641519.3657428, fang2024minisplattingrepresentingscenesconstrained}, thus the number of primitives should correlate to the level of detail describing the surface, which can be approximated as the number of pixels covering the scene.
Specifically, denoting the initial primitive number as $P_\mathrm{init}$, and the desired primitive number in the final 3DGS model as $P_\mathrm{fin}$, we control the primitive number of the $i$-th iteration by
\begin{equation}
    \label{eq:4}
    P_i = P_\mathrm{init} + \left(P_\mathrm{fin} - P_\mathrm{init}\right) / \left({r^{(i)}}\right)^{2 - i/S}, 
\end{equation}
where $r^{(i)}$ denotes the rendering resolution of the $i$-th iteration, powered by a factor of $2 - i/S$. 
In practice, we find too much growth of primitives by the end of optimization will result in poor convergence of the primitives, which significantly hinders the quality of the optimized 3DGS model. 
So we gradually decrease the power factor from 2 to 1 along the optimization process to suppress the primitive number in the early optimization while encouraging its growth in the mid-stage of optimization. Such a scheduling scheme produces a concave-up function depicting the relation between the primitive number and optimization steps with a small area under the curve (see \cref{fig:framework}), which largely reduces the optimization complexity. 

\paragraph{Momentum-based Primitive Budgeting.}
\vspace{-2mm}
Recent methods obtain $P_\mathrm{fin}$ relying on the dataset prior~\cite{fang2024minisplattingrepresentingscenesconstrained, mallick2024taming3dgshighqualityradiance, bulò2024revisingdensificationgaussiansplatting}, such as simply setting $P_\mathrm{fin}$ as a multiple of $P_\mathrm{init}$~\cite{mallick2024taming3dgshighqualityradiance}, which is a rule of thumb and cannot explain if it is enough for fitting a scene.
We address this problem alternatively by determining $P_\mathrm{fin}$ along the optimization process, instead of determining it ahead of optimization.

Specifically, inspired by the momentum optimizer~\cite{KingBa15}, which updates the parameters of a learning model with the accumulated gradients simulating the physical momentum, we regard $P_\mathrm{fin}$ as the momentum and the number of primitives naturally densified in each step as the force (denoted as $P_\mathrm{add}$). Then we calculate $P_\mathrm{fin}$ as 
\begin{equation}
    \label{eq:5}
    P_\mathrm{fin}^{(i)} = \max(P_\mathrm{fin}^{(i-1)},\thickspace \gamma\cdot P_\mathrm{fin}^{(i-1)} + \eta \cdot P_\mathrm{add}^{(i)}), 
\end{equation}
where $\gamma, \eta$ are hyper-parameters (defaulted by $0.98$ and $1$ in this paper), $i$ denotes the $i$-th optimization step.
The maximum operator is applied to keep $P_\mathrm{fin}^{(i)}$ monotonously increasing along the optimization process.
With \cref{eq:5}, we dynamically update $P_\mathrm{fin}$ in \cref{eq:4}, where $P_\mathrm{fin}^{(0)}$ is initialized as $5\cdot P_\mathrm{init}$. 
Experiments show that this momentum-based scheme performs effectively across various datasets, releasing densification scheduling from the need to specify an upper bound for the primitive growth.

\begin{table*}[t]
    \centering
    \footnotesize
    \setlength\tabcolsep{2pt}
    \captionsetup{singlelinecheck=false}
    \caption{
        Quantitative results of accelerating various 3DGS backbones using \OURS{}. 
        Plugging \OURS{} into different backbones, their optimization speed is largely improved by \ACC{} on average. 
        Most prominently, \OURS{} achieves acceleration with negligible compromise and even improvement in the rendering quality. 
        The results advocate that a reasonable computational resource allocation across the optimization process can largely boost the optimization while preserving the rendering quality. 
        ``Revising-3DGS*'' denotes our reproduction of the paper~\cite{bulò2024revisingdensificationgaussiansplatting} based on the Taming-3DGS~\cite{mallick2024taming3dgshighqualityradiance} codebase. 
        Time is reported in minutes. 
    }
    \begin{tabular} {l | rrrrr | rrrrr | rrrrr}
        \toprule
        \multirow{2}{*}{Method} 
        & \multicolumn{5}{c}{MipNeRF-360~\cite{Barron_2022_CVPR}} & \multicolumn{5}{c}{Deep Blending~\cite{DeepBlending2018}} & \multicolumn{5}{c}{Tanks and Temples~\cite{Knapitsch2017}} \\
        \cmidrule(l{2pt}r{2pt}){2-6} \cmidrule(l{2pt}r{2pt}){7-11} \cmidrule(l{2pt}r{2pt}){12-16}
        & $\mathrm{N_{GS}}\downarrow$ & PSNR$\uparrow$ & SSIM$\uparrow$ & LPIPS$\downarrow$ & Time$\downarrow$ 
        & $\mathrm{N_{GS}}\downarrow$ & PSNR$\uparrow$ & SSIM$\uparrow$ & LPIPS$\downarrow$ & Time$\downarrow$ 
        & $\mathrm{N_{GS}}\downarrow$ & PSNR$\uparrow$ & SSIM$\uparrow$ & LPIPS$\downarrow$ & Time$\downarrow$ \\
        \midrule
        3DGS~\cite{kerbl3Dgaussians}
                          &          3.12M &          27.72 &          0.826 & \textbf{0.201} &          18.31 
                          &          2.82M & \textbf{29.50} & \textbf{0.904} & \textbf{0.244} &          17.27 
                          &          1.83M &          23.62 & \textbf{0.847} & \textbf{0.177} &          10.59 \\
        +Ours
                          & \textbf{2.69M} & \textbf{27.81} & \textbf{0.828} &          0.203 & \textbf{10.16} 
                          & \textbf{2.18M} &          29.29 &          0.901 &          0.256 &  \textbf{8.16} 
                          & \textbf{1.43M} & \textbf{23.83} &          0.844 &          0.189 &  \textbf{6.13} \\
        \midrule
        Mip-Splatting~\cite{Yu2024MipSplatting}
                          &          3.97M &          27.91 & \textbf{0.838} & \textbf{0.174} &          25.83 
                          &          3.48M &          29.26 &          0.902 & \textbf{0.239} &          23.57 
                          &          2.36M &          23.75 & \textbf{0.859} & \textbf{0.156} &          14.23 \\
        +Ours
                          & \textbf{3.14M} & \textbf{27.99} &          0.835 &          0.186 & \textbf{12.60} 
                          & \textbf{2.55M} & \textbf{29.55} & \textbf{0.903} &          0.248 &  \textbf{9.80} 
                          & \textbf{1.62M} & \textbf{23.96} &          0.856 &          0.172 &  \textbf{6.13} \\
        \midrule
        Revising-3DGS*~\cite{bulò2024revisingdensificationgaussiansplatting}
                          &          1.94M &          27.65 &          0.826 &          0.194 &           5.73 
                          &          2.02M &          29.54 &          0.906 & \textbf{0.248} &           4.06 
                          &          1.32M & \textbf{23.87} & \textbf{0.856} & \textbf{0.168} &           4.50 \\
        +Ours
                          &          1.94M & \textbf{27.85} & \textbf{0.833} & \textbf{0.191} &  \textbf{3.46} 
                          &          2.02M & \textbf{29.76} & \textbf{0.907} &          0.251 &  \textbf{2.21} 
                          &          1.32M &          23.85 &          0.853 &          0.176 &  \textbf{3.16} \\
        \midrule
        Taming-3DGS~\cite{mallick2024taming3dgshighqualityradiance}
                          &          2.28M &          27.61 &          0.821 &          0.210 &           5.51 
                          &          2.32M &          29.69 &          0.906 & \textbf{0.245} &           4.52 
                          &          1.51M &          23.62 &          0.849 & \textbf{0.174} &           4.02 \\
        +Ours
                          & \textbf{2.04M} & \textbf{27.92} & \textbf{0.826} & \textbf{0.208} &  \textbf{3.23} 
                          & \textbf{1.94M} & \textbf{30.02} & \textbf{0.907} &          0.248 &  \textbf{2.20} 
                          & \textbf{1.19M} & \textbf{23.97} & \textbf{0.851} &          0.180 &  \textbf{2.62} \\
        \bottomrule
    \end{tabular}
    \label{tab:second-result}
\end{table*}

\section{Experiments}
\label{sec:exp}

\paragraph{Datasets.}
We conduct experiments on three real-world datasets, Mip-NeRF 360~\cite{Barron_2022_CVPR}, Deep-Blending~\cite{DeepBlending2018}, and Tanks\&Temples~\cite{Knapitsch2017}, respectively. 
For fair comparisons, we process these datasets following vanilla 3DGS~\cite{kerbl3Dgaussians}. 
We also perform evaluation on MatrixCity dataset~\cite{Li_2023_ICCV} for large-scale reconstruction in supplementary. 

\paragraph{Metrics.}
\vspace{-3mm}
To evaluate novel view rendering quality, we report average PSNR (peak signal-to-noise ratio) , SSIM~\cite{1284395}, and LPIPS~\cite{zhang2018perceptual} on each dataset. 
We also report the average total optimization time in minutes to measure the optimization speed. 
Furthermore, we also report the average primitive number of the optimized 3DGS model on each dataset, which gives an overall evaluation on per-primitive contribution to novel view rendering quality. 

\begin{figure}[t]
    \centering
    \captionsetup{singlelinecheck=false}
    \includegraphics[width=\linewidth]{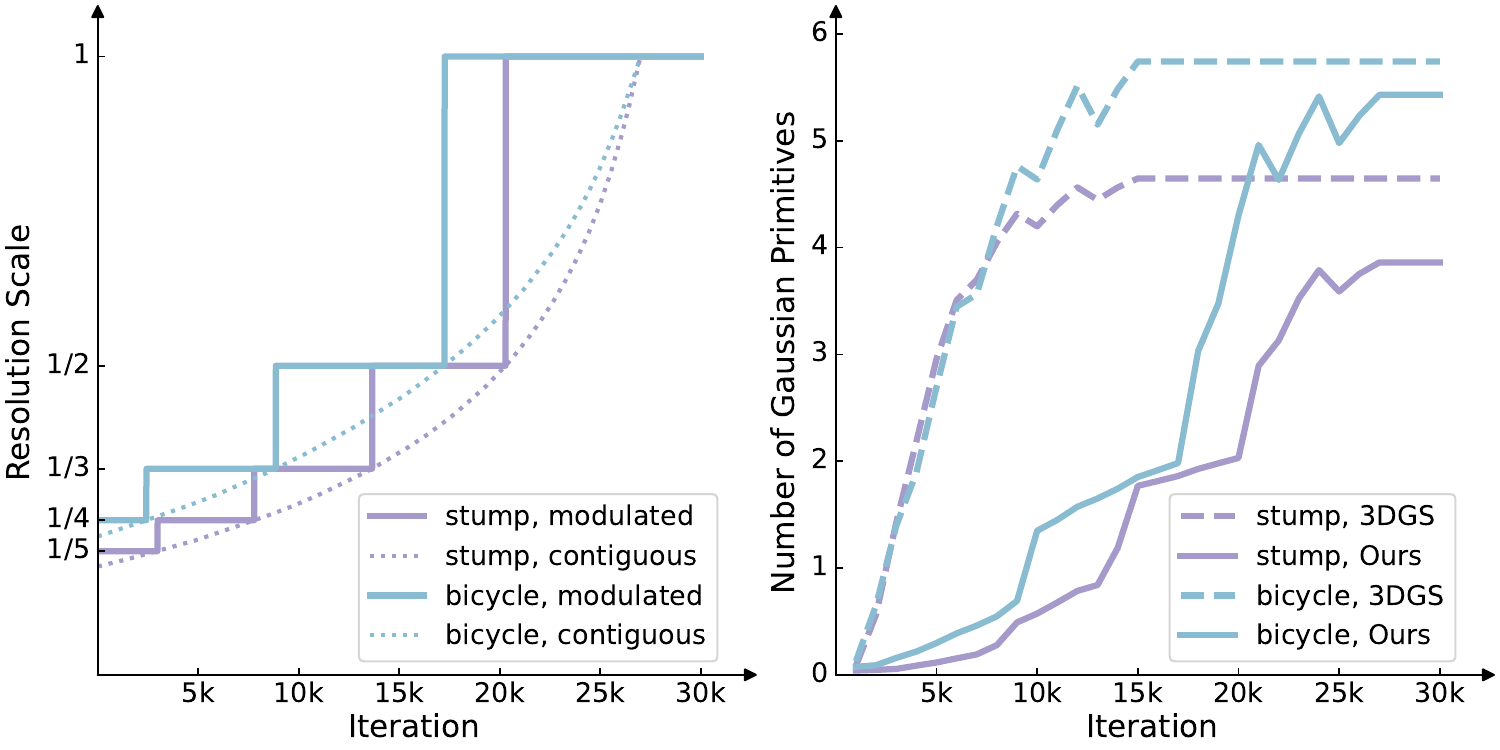}
    \vspace{-6mm}
    \caption{
        Variation in the optimization complexity. 
        The left shows that DashGaussian scene-adaptively schedules the rendering resolution for the scenes ``bicycle'' and ``stump'' from Mip-NeRF 360 dataset. 
        Please refer to implementation details (beginning of \cref{sec:exp}) for the definition of `contiguous' and `modulated'.
        The right shows the comparison on primitive growth between 3DGS~\cite{kerbl3Dgaussians} and 3DGS+DashGaussian on these two scenes. 
    }
    \label{fig:prim_and_reso}
\end{figure}

\paragraph{Implementation Details.}
\vspace{-3mm}
The resolution downsampling factor $r^{(i)}$ defined in \cref{sec:method:resosche} is a continuous scalar (see \cref{fig:framework}).
In practice, we modulate $r^{(i)}$ with $r^{(i)}=\lfloor r^{(i)} \rfloor$ (see \cref{fig:prim_and_reso}).
Such a modulation encourages the Gaussian primitives to grow relatively faster against the case without modulation, optimizing each Gaussian primitive more sufficiently to keep the novel view rendering quality. 

When we plug \OURS{} to backbones, the hyper-parameters of each backbone are kept unchanged as their official implementations.
All experiments are conducted on a single GPU with about 80 TFLOPS for FP32. 
Please refer to the supplementary for more implementation details. 

\subsection{Comparison with Fast Optimization Methods}
\label{sec:exp:sota}

\paragraph{Baselines.}
We report the comparisons with the state-of-the-art fast optimization methods, i.e., Reduced-3DGS~\cite{10.1145/3651282}, Mini-splatting~\cite{fang2024minisplattingrepresentingscenesconstrained} and Taming-3DGS~\cite{mallick2024taming3dgshighqualityradiance} in \cref{tab:main-result}.
Reduced-3DGS and Mini-Splatting achieve acceleration by pruning low contribution parameters, and Taming-3DGS obtains its speed for efficient back-propagation and Sparse Adam optimizer. 
These three methods are representative for algorithmic and engineering fast optimization methods. 
\blfootnote{$^1$Taming-3DGS~\cite{mallick2024taming3dgshighqualityradiance} referred to in this paper only includes efficient backward implementation and Sparse Adam optimizer.}

\paragraph{Quantitative Results.}
\vspace{-3mm}
By plugging \OURS{} to Taming-3DGS~\cite{mallick2024taming3dgshighqualityradiance}, we accomplish the optimization of 3DGS within 200 seconds, which is reported in \cref{tab:main-result}. 
Among all the methods in \cref{tab:main-result}, \OURS{} shows a significant improvement in the optimization speed over existing methods, and most importantly, with the SOTA rendering quality among these methods, which we attribute to the benefit from our reformulation of 3DGS optimization.

\paragraph{Qualitative Results.}
\vspace{-3mm}
We visualize the results of \cref{tab:main-result} in \cref{fig:expmainfigure}. 
Our method optimizes 3DGS within 200 seconds, while the rendering quality is still comparable to or even better than the baselines. 
The improvement comes from that \OURS{} effectively helps 3D Gaussians to better fit the captured observations from low to high frequency gradually with more reasonable densification. 

\subsection{\OURS{} Enhancing Backbones}
\label{sec:exp:enhance}

\paragraph{Quantitative Results.}
By equipping various 3DGS backbones with \OURS{}, we demonstrate that \OURS{} generalize well to 3DGS backbones with different optimizers~\cite{mallick2024taming3dgshighqualityradiance}, primitive representations~\cite{Yu2024MipSplatting}, and densification scores~\cite{bulò2024revisingdensificationgaussiansplatting}. 
Results are reported in \cref{tab:second-result}. 
The implementation details for each backbone are provided in the supplementary. 
Backbones enhanced with DashGaussian show a significant improvement in optimization speed compared with the vanilla ones, with an acceleration rate of \ACC{} on average. 
Besides, we also notice that \OURS{} requires fewer number of Gaussian primitives in fitting scenes. 
Despite of the significant acceleration and reduction in primitives, the enhanced backbones preserve their rendering quality compared to the vanilla ones. 
The results demonstrate the effectiveness and generalization of our method, supporting \OURS{} as a plug-and-play optimization strategy that is able to be integrated into any other 3DGS baselines rather than a dedicated improvement on a specific backbone. 

\begin{figure}[t]
    \centering
    \captionsetup{singlelinecheck=false}
    \includegraphics[width=0.9\linewidth]{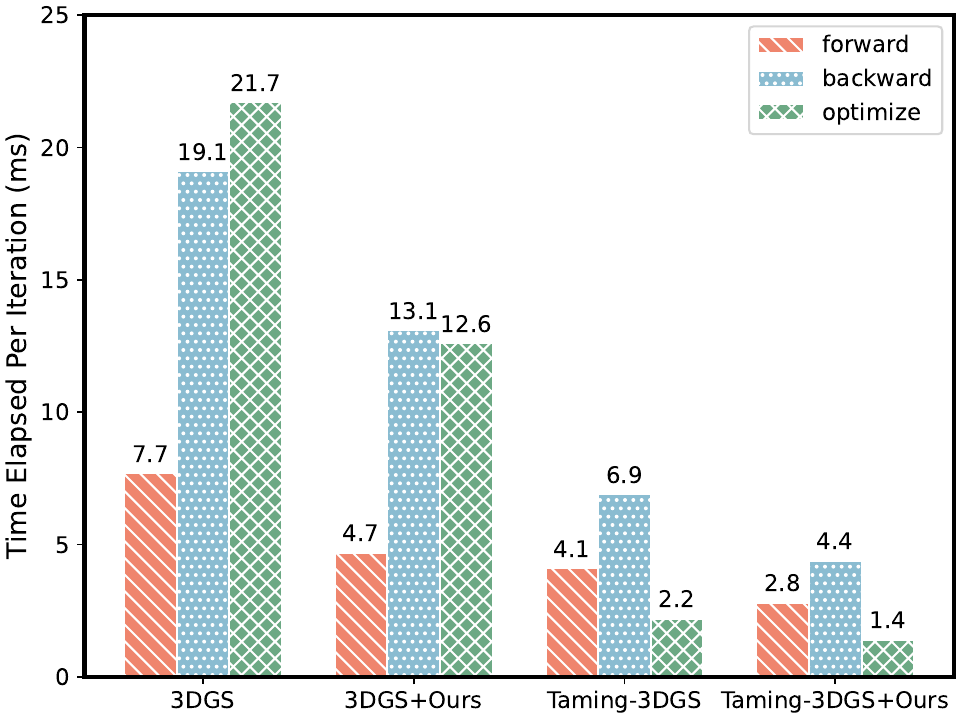}
    \vspace{-2mm}
    \caption{
        Profiling of the average optimization time on the Mip-NeRF 360 dataset. 
        We can see \OURS{} effectively reduces the per-iteration elapsed time of all three major operations. 
        It verifies that \OURS{} can effectively save the overall computational complexity of 3DGS optimization. 
    }
    \vspace{-4mm}
    \label{fig:profiling}
\end{figure}

\paragraph{Visualization of Schedulers. }
\vspace{-4mm}
We show the resolution scheduler and the Gaussian primitive scheduler in two different scenes (``bicycle'' and ``stump'') from the Mip-NeRF 360 dataset in \cref{fig:prim_and_reso}. 
Specifically, on the left of \cref{fig:prim_and_reso}, we compare our resolution scheduler in different scenes. 
Images in different scenes manifest diverse frequency information, and our method can scene-adaptively schedule the rendering resolution. 
On the right of \cref{fig:prim_and_reso}, we compare the primitive growth between 3DGS~\cite{kerbl3Dgaussians} and its scheduled version (marked as ``Ours''). 
Noticeably, our method represents scenes with fewer primitives. 
Please check the supplementary for more analysis on this phenomenon. 

\paragraph{Profiling of the Optimization Time.}
\vspace{-4mm}
As we pointed out in \cref{sec:intro}, the three most time-consuming operations in one optimization step of 3DGS are forward-rendering, gradient back-propagation, and primitive optimizing, respectively. 
We profile the average time cost of them over the whole optimization process on 3DGS~\cite{kerbl3Dgaussians} and Taming-3DGS~\cite{mallick2024taming3dgshighqualityradiance} to see how \OURS{} affects these two backbones with different optimizer.
The result is shown in \cref{fig:profiling}.
\OURS{} brings significant reduction on all of the three components, while the acceleration rate is almost consistent. 
It demonstrates that \OURS{} can substantially speed up the optimization regard less of different implementations of the 3DGS backbone.  

\vspace{1mm}
\subsection{Ablation Study}
\label{sec:exp:ablation}
We use Taming-3DGS~\cite{mallick2024taming3dgshighqualityradiance} as the backbone and individually add each module in \OURS{} to explore their effect on the rendering quality and optimization speed.

\paragraph{Resolution Scheduler.}
\vspace{-3mm}
We keep the resolution scheduler proposed in \cref{sec:method:primsche}, and remove the primitive scheduler to evaluate the resolution scheduler. 
The corresponding result in \cref{tab:ablation} is marked as ``+Reso-Sche.'', which exhibits an allround improvement over the backbone. 

\paragraph{Primitive Scheduler.}
\vspace{-3mm}
To evaluate the primitive scheduler in \cref{sec:method:primsche}, we have the primitive schedule depending on the resolution schedule while disabling the downsampling operation on training views to supervise 3DGS with the original resolution. 
The corresponding result in \cref{tab:ablation} is marked as ``+Prim-Sche.''. 
Acceleration and quality improvement are similarly observed as ``+Reso-Sche.''.

\paragraph{Full Method.}
\vspace{-3mm}
We present the result of equipping the backbone with the full \OURS{}, reported as ``Full'' in \cref{tab:ablation}.
By having the two schedulers cooperating with each other, \OURS{} significantly reduces the optimization time of Taming-3DGS by 41.3\%.
Meanwhile, our method surpasses the rendering quality of the backbone while using fewer Gaussian primitives. 

\paragraph{Hyper-parameters.}
\vspace{-3mm}
The hyper-parameters of \OURS{} can influence the optimization time and rendering quality. 
We report the ablation studies on them in the supplementary, especially $\gamma$ and $\eta$ in the momentum-based primitive budgeting method that potentially influence $P_\mathrm{fin}$. 

\begin{table}[t]
    \centering
    \captionsetup{singlelinecheck=false}
    \caption{
        Ablation studies over the proposed scheduling schemes of DashGaussian. 
        Experiments are performed on the Mip-NeRF 360 dataset~\cite{Barron_2022_CVPR} with Taming-3DGS~\cite{mallick2024taming3dgshighqualityradiance} as the backbone. 
    }
    \resizebox{\linewidth}{!}{
    
        \begin{tabular} {l | rrrrr}
            \toprule
            Method                    & $N_\mathrm{GS}\downarrow$ & PSNR$\uparrow$ & SSIM$\uparrow$ & LPIPS$\downarrow$ & Time$\downarrow$ \\
            \midrule
            Taming-3DGS               &            2.28M &          27.61 &          0.821 &             0.210 &             5.51 \\
            +Reso-Sche.               &            2.46M & \textbf{27.99} & \textbf{0.829} &    \textbf{0.196} &             4.43 \\
            +Prim-Sche.               &            2.23M &          27.85 &          0.824 &             0.208 &             4.60 \\
            \midrule
            Full                      &   \textbf{2.04M} &          27.92 &          0.826 &             0.208 &    \textbf{3.23} \\
            \bottomrule
        \end{tabular}
    }
    \label{tab:ablation}
\end{table}
\begin{table}
    \centering
    \captionsetup{singlelinecheck=false}
    \footnotesize
    \caption{
        We report the primitive number $N_\mathrm{GS}^\mathrm{bb}$ of the backbones, the primitive number $P_\mathrm{fin}$ predicted by our momentum-based budgeting, and the primitive number $N_\mathrm{GS}^\mathrm{ours}$ of DashGaussian-scheduled backbones. 
        Conducted on Mip-NeRF 360 dataset. 
    }
    \begin{tabular} {l|rrr}
        \toprule
        Method    & $N_\mathrm{GS}^\mathrm{bb}$ & $P_\mathrm{fin}$ & $N_\mathrm{GS}^\mathrm{ours}$\\
        \midrule
        3DGS~\cite{kerbl3Dgaussians}
                  & 3.12M & 3.08M & 2.69M \\
        Mip-Splatting~\cite{Yu2024MipSplatting}
                  & 3.97M & 3.25M & 3.14M \\
        Taming-3DGS~\cite{mallick2024taming3dgshighqualityradiance}
                  & 2.28M & 2.39M & 2.04M \\
        \bottomrule
    \end{tabular}
    \label{tab:momentum}
\end{table}

\subsection{Primitive Growth Upperbound}
\label{sec:exp:primitive_upperbound}

We evaluate how precise our momentum-based primitive budget method can estimate the maximum primitive number of a naturally densified 3DGS. 
The results are reported in \cref{tab:momentum}. 
Our method effectively estimates an approximate number of primitives to densify for different backbones, while we have $N_\mathrm{GS}^\mathrm{ours}$ less than $P_\mathrm{fin}$ because there are no more primitives meeting the densification condition of the backbones.
Further experiments in the supplementary demonstrate that the primitive number in the optimized \OURS{} tends to be a constant for the same scene and will not increase when $P_\mathrm{fin}$ is extremely large, which prevents \OURS{} from out of memory.
\vspace{0mm}
\section{Conclusion}
\label{sec:conclusion}

In this paper, we propose \OURS{}, a computational resource allocation method to accelerate the optimization of 3DGS. 
Our method innovatively formulates 3DGS optimization as a frequency component fitting process, allocates the rendering resolution and primitive number corporately, and boosts the optimization by a large margin without compromise in rendering quality. 
Our method can be easily plugged into various 3DGS backbones, possessing the potential as a new optimization paradigm for 3DGS. 
This work can also improve the feasibility of training 3DGS on large-scale, time-intensive reconstruction tasks. 
In the future work, we will consider integrating a more insightful analysis into the connection between the primitive number and rendering resolution as a possible direction.

{
    \small
    \bibliographystyle{ieeenat_fullname}
    \bibliography{main}

\begin{thebibliography}{50}
\providecommand{\natexlab}[1]{#1}
\providecommand{\url}[1]{\texttt{#1}}
\expandafter\ifx\csname urlstyle\endcsname\relax
  \providecommand{\doi}[1]{doi: #1}\else
  \providecommand{\doi}{doi: \begingroup \urlstyle{rm}\Url}\fi

\bibitem[Aliev et~al.(2020)Aliev, Sevastopolsky, Kolos, Ulyanov, and Lempitsky]{10.1007/978-3-030-58542-6_42}
Kara-Ali Aliev, Artem Sevastopolsky, Maria Kolos, Dmitry Ulyanov, and Victor Lempitsky.
\newblock Neural point-based graphics.
\newblock In \emph{Computer Vision -- ECCV 2020}, pages 696--712, Cham, 2020. Springer International Publishing.

\bibitem[Barron et~al.(2021)Barron, Mildenhall, Tancik, Hedman, Martin-Brualla, and Srinivasan]{Barron_2021_ICCV}
Jonathan~T. Barron, Ben Mildenhall, Matthew Tancik, Peter Hedman, Ricardo Martin-Brualla, and Pratul~P. Srinivasan.
\newblock Mip-nerf: A multiscale representation for anti-aliasing neural radiance fields.
\newblock In \emph{Proceedings of the IEEE/CVF International Conference on Computer Vision (ICCV)}, pages 5855--5864, 2021.

\bibitem[Barron et~al.(2022)Barron, Mildenhall, Verbin, Srinivasan, and Hedman]{Barron_2022_CVPR}
Jonathan~T. Barron, Ben Mildenhall, Dor Verbin, Pratul~P. Srinivasan, and Peter Hedman.
\newblock Mip-nerf 360: Unbounded anti-aliased neural radiance fields.
\newblock In \emph{Proceedings of the IEEE/CVF Conference on Computer Vision and Pattern Recognition (CVPR)}, pages 5470--5479, 2022.

\bibitem[Buehler et~al.(2001)Buehler, Bosse, McMillan, Gortler, and Cohen]{buehler2001unstructured}
Chris Buehler, Michael Bosse, Leonard McMillan, Steven Gortler, and Michael Cohen.
\newblock Unstructured lumigraph rendering.
\newblock In \emph{Proceedings of the 28th annual conference on Computer graphics and interactive techniques}, pages 425--432, 2001.

\bibitem[Bulò et~al.(2024)Bulò, Porzi, and Kontschieder]{bulò2024revisingdensificationgaussiansplatting}
Samuel~Rota Bulò, Lorenzo Porzi, and Peter Kontschieder.
\newblock Revising densification in gaussian splatting, 2024.

\bibitem[Chen et~al.(2022)Chen, Xu, Geiger, Yu, and Su]{10.1007/978-3-031-19824-3_20}
Anpei Chen, Zexiang Xu, Andreas Geiger, Jingyi Yu, and Hao Su.
\newblock Tensorf: Tensorial radiance fields.
\newblock In \emph{Computer Vision -- ECCV 2022}, pages 333--350, Cham, 2022. Springer Nature Switzerland.

\bibitem[Chen et~al.(2024)Chen, Nie, Ummenhofer, Birkl, Paulitsch, Müller, and Nießner]{chen2024mesh2nerfdirectmeshsupervision}
Yujin Chen, Yinyu Nie, Benjamin Ummenhofer, Reiner Birkl, Michael Paulitsch, Matthias Müller, and Matthias Nießner.
\newblock Mesh2nerf: Direct mesh supervision for neural radiance field representation and generation, 2024.

\bibitem[Chen et~al.(2023)Chen, Funkhouser, Hedman, and Tagliasacchi]{Chen_2023_CVPR}
Zhiqin Chen, Thomas Funkhouser, Peter Hedman, and Andrea Tagliasacchi.
\newblock Mobilenerf: Exploiting the polygon rasterization pipeline for efficient neural field rendering on mobile architectures.
\newblock In \emph{Proceedings of the IEEE/CVF Conference on Computer Vision and Pattern Recognition (CVPR)}, pages 16569--16578, 2023.

\bibitem[Debevec(2008)]{10.1145/1401132.1401175}
Paul Debevec.
\newblock Rendering synthetic objects into real scenes: bridging traditional and image-based graphics with global illumination and high dynamic range photography.
\newblock In \emph{ACM SIGGRAPH 2008 Classes}, New York, NY, USA, 2008. Association for Computing Machinery.

\bibitem[Deng et~al.(2025)Deng, Diao, Li, Yu, and Xu]{deng2025efficientdensitycontrol3d}
Xiaobin Deng, Changyu Diao, Min Li, Ruohan Yu, and Duanqing Xu.
\newblock Efficient density control for 3d gaussian splatting, 2025.

\bibitem[Duckworth et~al.(2024)Duckworth, Hedman, Reiser, Zhizhin, Thibert, Lu\v{c}i\'{c}, Szeliski, and Barron]{10.1145/3658193}
Daniel Duckworth, Peter Hedman, Christian Reiser, Peter Zhizhin, Jean-Fran\c{c}ois Thibert, Mario Lu\v{c}i\'{c}, Richard Szeliski, and Jonathan~T. Barron.
\newblock Smerf: Streamable memory efficient radiance fields for real-time large-scene exploration.
\newblock 43\penalty0 (4), 2024.

\bibitem[Durvasula et~al.(2023)Durvasula, Zhao, Chen, Liang, Sanjaya, and Vijaykumar]{durvasula2023distwarfastdifferentiablerendering}
Sankeerth Durvasula, Adrian Zhao, Fan Chen, Ruofan Liang, Pawan~Kumar Sanjaya, and Nandita Vijaykumar.
\newblock Distwar: Fast differentiable rendering on raster-based rendering pipelines, 2023.

\bibitem[Fang and Wang(2024)]{fang2024minisplattingrepresentingscenesconstrained}
Guangchi Fang and Bing Wang.
\newblock Mini-splatting: Representing scenes with a constrained number of gaussians, 2024.

\bibitem[Feng et~al.(2024)Feng, Chen, Fu, Liao, Wang, Liu, Pei, Li, Zhang, and Dai]{feng2024flashgsefficient3dgaussian}
Guofeng Feng, Siyan Chen, Rong Fu, Zimu Liao, Yi Wang, Tao Liu, Zhilin Pei, Hengjie Li, Xingcheng Zhang, and Bo Dai.
\newblock Flashgs: Efficient 3d gaussian splatting for large-scale and high-resolution rendering, 2024.

\bibitem[Fitzgibbon et~al.(2005)Fitzgibbon, Wexler, and Zisserman]{fitzgibbon2005image}
Andrew Fitzgibbon, Yonatan Wexler, and Andrew Zisserman.
\newblock Image-based rendering using image-based priors.
\newblock \emph{International Journal of Computer Vision}, 63:\penalty0 141--151, 2005.

\bibitem[Fridovich-Keil et~al.(2022)Fridovich-Keil, Yu, Tancik, Chen, Recht, and Kanazawa]{Fridovich-Keil_2022_CVPR}
Sara Fridovich-Keil, Alex Yu, Matthew Tancik, Qinhong Chen, Benjamin Recht, and Angjoo Kanazawa.
\newblock Plenoxels: Radiance fields without neural networks.
\newblock In \emph{Proceedings of the IEEE/CVF Conference on Computer Vision and Pattern Recognition (CVPR)}, pages 5501--5510, 2022.

\bibitem[Gonzalez and Woods(2006)]{10.5555/1076432}
Rafael~C. Gonzalez and Richard~E. Woods.
\newblock \emph{Digital Image Processing (3rd Edition)}.
\newblock Prentice-Hall, Inc., USA, 2006.

\bibitem[Hedman et~al.(2018)Hedman, Philip, Price, Frahm, Drettakis, and Brostow]{DeepBlending2018}
Peter Hedman, Julien Philip, True Price, Jan-Michael Frahm, George Drettakis, and Gabriel Brostow.
\newblock Deep blending for free-viewpoint image-based rendering.
\newblock 37\penalty0 (6):\penalty0 257:1--257:15, 2018.

\bibitem[Hu et~al.(2024)Hu, Wang, Fan, Fan, Peng, Lei, Li, and Zhang]{hu2024sagdboundaryenhancedsegment3d}
Xu Hu, Yuxi Wang, Lue Fan, Junsong Fan, Junran Peng, Zhen Lei, Qing Li, and Zhaoxiang Zhang.
\newblock Sagd: Boundary-enhanced segment anything in 3d gaussian via gaussian decomposition, 2024.

\bibitem[Huang et~al.(2024)Huang, Yu, Chen, Geiger, and Gao]{10.1145/3641519.3657428}
Binbin Huang, Zehao Yu, Anpei Chen, Andreas Geiger, and Shenghua Gao.
\newblock 2d gaussian splatting for geometrically accurate radiance fields.
\newblock In \emph{ACM SIGGRAPH 2024 Conference Papers}, New York, NY, USA, 2024. Association for Computing Machinery.

\bibitem[Kerbl et~al.(2023)Kerbl, Kopanas, Leimk{\"u}hler, and Drettakis]{kerbl3Dgaussians}
Bernhard Kerbl, Georgios Kopanas, Thomas Leimk{\"u}hler, and George Drettakis.
\newblock 3d gaussian splatting for real-time radiance field rendering.
\newblock \emph{ACM Transactions on Graphics}, 42\penalty0 (4), 2023.

\bibitem[Kheradmand et~al.(2024)Kheradmand, Rebain, Sharma, Sun, Tseng, Isack, Kar, Tagliasacchi, and Yi]{NEURIPS2024_93be245f}
Shakiba Kheradmand, Daniel Rebain, Gopal Sharma, Weiwei Sun, Yang-Che Tseng, Hossam Isack, Abhishek Kar, Andrea Tagliasacchi, and Kwang~Moo Yi.
\newblock 3d gaussian splatting as markov chain monte carlo.
\newblock In \emph{Advances in Neural Information Processing Systems}, pages 80965--80986. Curran Associates, Inc., 2024.

\bibitem[Kingma and Ba(2015)]{KingBa15}
Diederik Kingma and Jimmy Ba.
\newblock Adam: A method for stochastic optimization.
\newblock In \emph{International Conference on Learning Representations (ICLR)}, San Diega, CA, USA, 2015.

\bibitem[Knapitsch et~al.(2017)Knapitsch, Park, Zhou, and Koltun]{Knapitsch2017}
Arno Knapitsch, Jaesik Park, Qian-Yi Zhou, and Vladlen Koltun.
\newblock Tanks and temples: Benchmarking large-scale scene reconstruction.
\newblock \emph{ACM Transactions on Graphics}, 36\penalty0 (4), 2017.

\bibitem[Li et~al.(2025)Li, Zhang, Bai, Zheng, Ning, Zhou, and Gu]{10.1007/978-3-031-72684-2_8}
Jiahe Li, Jiawei Zhang, Xiao Bai, Jin Zheng, Xin Ning, Jun Zhou, and Lin Gu.
\newblock Talkinggaussian: Structure-persistent 3d talking head synthesis via gaussian splatting.
\newblock In \emph{Computer Vision -- ECCV 2024}, pages 127--145, Cham, 2025. Springer Nature Switzerland.

\bibitem[Li et~al.(2023)Li, Jiang, Xu, Xiangli, Wang, Lin, and Dai]{Li_2023_ICCV}
Yixuan Li, Lihan Jiang, Linning Xu, Yuanbo Xiangli, Zhenzhi Wang, Dahua Lin, and Bo Dai.
\newblock Matrixcity: A large-scale city dataset for city-scale neural rendering and beyond.
\newblock In \emph{Proceedings of the IEEE/CVF International Conference on Computer Vision (ICCV)}, pages 3205--3215, 2023.

\bibitem[Liang et~al.(2025)Liang, Zhang, Hu, Zhu, Feng, and Jia]{10.1007/978-3-031-72643-9_17}
Zhihao Liang, Qi Zhang, Wenbo Hu, Lei Zhu, Ying Feng, and Kui Jia.
\newblock Analytic-splatting: Anti-aliased 3d gaussian splatting via analytic integration.
\newblock In \emph{Computer Vision -- ECCV 2024}, pages 281--297, Cham, 2025. Springer Nature Switzerland.

\bibitem[Lin et~al.(2024)Lin, Li, Tang, Liu, Liu, Liu, Lu, Wu, Xu, Yan, and Yang]{Lin_2024_CVPR}
Jiaqi Lin, Zhihao Li, Xiao Tang, Jianzhuang Liu, Shiyong Liu, Jiayue Liu, Yangdi Lu, Xiaofei Wu, Songcen Xu, Youliang Yan, and Wenming Yang.
\newblock Vastgaussian: Vast 3d gaussians for large scene reconstruction.
\newblock In \emph{Proceedings of the IEEE/CVF Conference on Computer Vision and Pattern Recognition (CVPR)}, pages 5166--5175, 2024.

\bibitem[Liu et~al.(2020)Liu, Gu, Lin, Chua, and Theobalt]{liu2020neural}
Lingjie Liu, Jiatao Gu, Kyaw~Zaw Lin, Tat-Seng Chua, and Christian Theobalt.
\newblock Neural sparse voxel fields.
\newblock \emph{NeurIPS}, 2020.

\bibitem[Liu et~al.(2024)Liu, Guan, Luo, Fan, Peng, and Zhang]{liu2024citygaussian}
Yang Liu, He Guan, Chuanchen Luo, Lue Fan, Junran Peng, and Zhaoxiang Zhang.
\newblock Citygaussian: Real-time high-quality large-scale scene rendering with gaussians, 2024.

\bibitem[Mallick et~al.(2024)Mallick, Goel, Kerbl, Carrasco, Steinberger, and Torre]{mallick2024taming3dgshighqualityradiance}
Saswat~Subhajyoti Mallick, Rahul Goel, Bernhard Kerbl, Francisco~Vicente Carrasco, Markus Steinberger, and Fernando De~La Torre.
\newblock Taming 3dgs: High-quality radiance fields with limited resources, 2024.

\bibitem[Mildenhall et~al.(2021)Mildenhall, Srinivasan, Tancik, Barron, Ramamoorthi, and Ng]{10.1145/3503250}
Ben Mildenhall, Pratul~P. Srinivasan, Matthew Tancik, Jonathan~T. Barron, Ravi Ramamoorthi, and Ren Ng.
\newblock Nerf: representing scenes as neural radiance fields for view synthesis.
\newblock \emph{Commun. ACM}, 65\penalty0 (1):\penalty0 99–106, 2021.

\bibitem[M\"uller et~al.(2022)M\"uller, Evans, Schied, and Keller]{mueller2022instant}
Thomas M\"uller, Alex Evans, Christoph Schied, and Alexander Keller.
\newblock Instant neural graphics primitives with a multiresolution hash encoding.
\newblock \emph{ACM Trans. Graph.}, 41\penalty0 (4):\penalty0 102:1--102:15, 2022.

\bibitem[Papantonakis et~al.(2024)Papantonakis, Kopanas, Kerbl, Lanvin, and Drettakis]{10.1145/3651282}
Panagiotis Papantonakis, Georgios Kopanas, Bernhard Kerbl, Alexandre Lanvin, and George Drettakis.
\newblock Reducing the memory footprint of 3d gaussian splatting.
\newblock 7\penalty0 (1), 2024.

\bibitem[Remondino and El-Hakim(2006)]{remondino2006image}
Fabio Remondino and Sabry El-Hakim.
\newblock Image-based 3d modelling: a review.
\newblock \emph{The photogrammetric record}, 21\penalty0 (115):\penalty0 269--291, 2006.

\bibitem[Schonberger and Frahm(2016)]{Schonberger_2016_CVPR}
Johannes~L. Schonberger and Jan-Michael Frahm.
\newblock Structure-from-motion revisited.
\newblock In \emph{Proceedings of the IEEE Conference on Computer Vision and Pattern Recognition (CVPR)}, 2016.

\bibitem[Seitz and Dyer(1999)]{seitz1999photorealistic}
Steven~M Seitz and Charles~R Dyer.
\newblock Photorealistic scene reconstruction by voxel coloring.
\newblock \emph{International journal of computer vision}, 35:\penalty0 151--173, 1999.

\bibitem[Seo et~al.(2024)Seo, Choi, Son, and Uh]{seo2024flodintegratingflexiblelevel}
Yunji Seo, Young~Sun Choi, Hyun~Seung Son, and Youngjung Uh.
\newblock Flod: Integrating flexible level of detail into 3d gaussian splatting for customizable rendering, 2024.

\bibitem[Shen et~al.(2025)Shen, Yang, and Wang]{10.1007/978-3-031-72670-5_26}
Qiuhong Shen, Xingyi Yang, and Xinchao Wang.
\newblock Flashsplat: 2d to 3d gaussian splatting segmentation solved optimally.
\newblock In \emph{Computer Vision -- ECCV 2024}, pages 456--472, Cham, 2025. Springer Nature Switzerland.

\bibitem[Shi et~al.(2024)Shi, Gasparini, Morin, and Ooi]{shi2024lapisgslayeredprogressive3d}
Yuang Shi, Simone Gasparini, Géraldine Morin, and Wei~Tsang Ooi.
\newblock Lapisgs: Layered progressive 3d gaussian splatting for adaptive streaming, 2024.

\bibitem[Tewari et~al.(2022)Tewari, Thies, Mildenhall, Srinivasan, Tretschk, Yifan, Lassner, Sitzmann, Martin-Brualla, Lombardi, Simon, Theobalt, Nießner, Barron, Wetzstein, Zollhöfer, and Golyanik]{https://doi.org/10.1111/cgf.14507}
A. Tewari, J. Thies, B. Mildenhall, P. Srinivasan, E. Tretschk, W. Yifan, C. Lassner, V. Sitzmann, R. Martin-Brualla, S. Lombardi, T. Simon, C. Theobalt, M. Nießner, J.~T. Barron, G. Wetzstein, M. Zollhöfer, and V. Golyanik.
\newblock Advances in neural rendering.
\newblock \emph{Computer Graphics Forum}, 41\penalty0 (2):\penalty0 703--735, 2022.

\bibitem[Wang et~al.(2004)Wang, Bovik, Sheikh, and Simoncelli]{1284395}
Zhou Wang, A.C. Bovik, H.R. Sheikh, and E.P. Simoncelli.
\newblock Image quality assessment: from error visibility to structural similarity.
\newblock \emph{IEEE Transactions on Image Processing}, 13\penalty0 (4):\penalty0 600--612, 2004.

\bibitem[Wang et~al.(2023)Wang, Shen, Nimier-David, Sharp, Gao, Keller, Fidler, M\"uller, and Gojcic]{adaptiveshells2023}
Zian Wang, Tianchang Shen, Merlin Nimier-David, Nicholas Sharp, Jun Gao, Alexander Keller, Sanja Fidler, Thomas M\"uller, and Zan Gojcic.
\newblock Adaptive shells for efficient neural radiance field rendering.
\newblock \emph{ACM Trans. Graph.}, 42\penalty0 (6), 2023.

\bibitem[Ye et~al.(2024)Ye, Li, Kerr, Turkulainen, Yi, Pan, Seiskari, Ye, Hu, Tancik, and Kanazawa]{ye2024gsplatopensourcelibrarygaussian}
Vickie Ye, Ruilong Li, Justin Kerr, Matias Turkulainen, Brent Yi, Zhuoyang Pan, Otto Seiskari, Jianbo Ye, Jeffrey Hu, Matthew Tancik, and Angjoo Kanazawa.
\newblock gsplat: An open-source library for gaussian splatting, 2024.

\bibitem[Yu et~al.(2024{\natexlab{a}})Yu, Qu, Yu, Chen, Jiang, Chen, Zhang, Xu, Wu, Lv, and Yu]{10.1145/3664647.3681675}
Hongyun Yu, Zhan Qu, Qihang Yu, Jianchuan Chen, Zhonghua Jiang, Zhiwen Chen, Shengyu Zhang, Jimin Xu, Fei Wu, Chengfei Lv, and Gang Yu.
\newblock Gaussiantalker: Speaker-specific talking head synthesis via 3d gaussian splatting.
\newblock page 3548–3557, New York, NY, USA, 2024{\natexlab{a}}. Association for Computing Machinery.

\bibitem[Yu et~al.(2024{\natexlab{b}})Yu, Chen, Huang, Sattler, and Geiger]{Yu2024MipSplatting}
Zehao Yu, Anpei Chen, Binbin Huang, Torsten Sattler, and Andreas Geiger.
\newblock Mip-splatting: Alias-free 3d gaussian splatting.
\newblock In \emph{Proceedings of the IEEE/CVF Conference on Computer Vision and Pattern Recognition (CVPR)}, pages 19447--19456, 2024{\natexlab{b}}.

\bibitem[Yu et~al.(2024{\natexlab{c}})Yu, Sattler, and Geiger]{Yu2024GOF}
Zehao Yu, Torsten Sattler, and Andreas Geiger.
\newblock Gaussian opacity fields: Efficient adaptive surface reconstruction in unbounded scenes.
\newblock \emph{ACM Transactions on Graphics}, 2024{\natexlab{c}}.

\bibitem[Yu~Chen(2024)]{yuchen2024dogaussian}
Gim Hee~Lee Yu~Chen.
\newblock Dogs: Distributed-oriented gaussian splatting for large-scale 3d reconstruction via gaussian consensus.
\newblock In \emph{arXiv}, 2024.

\bibitem[Zhang et~al.(2018)Zhang, Isola, Efros, Shechtman, and Wang]{zhang2018perceptual}
Richard Zhang, Phillip Isola, Alexei~A Efros, Eli Shechtman, and Oliver Wang.
\newblock The unreasonable effectiveness of deep features as a perceptual metric.
\newblock In \emph{CVPR}, 2018.

\bibitem[Zhong et~al.(2024)Zhong, Hong, Li, and Xu]{zhong2024cvtxrfcontrastiveinvoxeltransformer}
Yingji Zhong, Lanqing Hong, Zhenguo Li, and Dan Xu.
\newblock Cvt-xrf: Contrastive in-voxel transformer for 3d consistent radiance fields from sparse inputs, 2024.

\end{thebibliography}
}

\clearpage
\maketitlesupplementary

\setcounter{section}{6}
\setcounter{table}{4}
\setcounter{figure}{5}
\setcounter{equation}{5}

\noindent In this supplementary material, we include explanations of the implementation details in \cref{sup:details}, the discussion over primitive growth in \cref{sup:primgrowth} and furthermore extensive experiments in \cref{sup:scene-wise}, \cref{sup:ablation-param} and \cref{sup:largescale}.


\begin{table*}[t]
    \centering
    \footnotesize
    \setlength\tabcolsep{2pt}
    \captionsetup{singlelinecheck=false}
    \caption{
        Scene-wise quantitative results over Mip-NeRF 360 dataset~\cite{Barron_2022_CVPR}. 
    }
    \begin{tabular} {l | rrrrr | rrrrr | rrrrr}
        \toprule
        \multirow{2}{*}{Method} 
        & \multicolumn{5}{c}{bicycle} & \multicolumn{5}{c}{garden} & \multicolumn{5}{c}{stump} \\
        \cmidrule(l{2pt}r{2pt}){2-6} \cmidrule(l{2pt}r{2pt}){7-11} \cmidrule(l{2pt}r{2pt}){12-16}
        & $\mathrm{N_{GS}}\downarrow$ & PSNR$\uparrow$ & SSIM$\uparrow$ & LPIPS$\downarrow$ & Time$\downarrow$ 
        & $\mathrm{N_{GS}}\downarrow$ & PSNR$\uparrow$ & SSIM$\uparrow$ & LPIPS$\downarrow$ & Time$\downarrow$ 
        & $\mathrm{N_{GS}}\downarrow$ & PSNR$\uparrow$ & SSIM$\uparrow$ & LPIPS$\downarrow$ & Time$\downarrow$ \\
        \midrule
        3DGS~\cite{kerbl3Dgaussians}
                          &           5.74M &          25.61 &          0.778 &          0.203 &          24.82 
                          &           5.10M &          27.80 & \textbf{0.874} & \textbf{0.102} &          23.99 
                          &           4.64M &          26.91 &          0.784 &          0.207 &          19.87 \\
        +Ours
                          &  \textbf{5.38M} & \textbf{25.70} & \textbf{0.785} & \textbf{0.202} & \textbf{14.86} 
                          &  \textbf{3.42M} & \textbf{27.93} &          0.873 &          0.110 & \textbf{11.72} 
                          &  \textbf{3.80M} & \textbf{27.32} & \textbf{0.797} & \textbf{0.201} &  \textbf{9.64} \\
        \midrule
        Mip-Splatting~\cite{Yu2024MipSplatting}
                          &           7.82M & \textbf{25.95} & \textbf{0.803} & \textbf{0.162} &          38.81 
                          &           5.60M &          27.90 & \textbf{0.884} & \textbf{0.089} &          32.78 
                          &           5.66M &          27.15 & \textbf{0.800} & \textbf{0.181} &          28.57 \\
        +Ours
                          &  \textbf{5.42M} &          25.85 &          0.796 &          0.176 & \textbf{17.67} 
                          &  \textbf{3.87M} & \textbf{28.18} &          0.882 &          0.096 & \textbf{14.89} 
                          &  \textbf{4.58M} & \textbf{27.20} &          0.800 &          0.189 & \textbf{11.39} \\
        \midrule
        Revising-3DGS*~\cite{bulò2024revisingdensificationgaussiansplatting}
                          &           3.20M &          25.58 &          0.776 &          0.205 &           5.44 
                          &           2.65M &          27.75 &          0.873 &          0.106 &           6.78 
                          &           3.03M &          27.04 &          0.797 &          0.191 &           5.43 \\
        +Ours
                          &           3.20M & \textbf{25.66} & \textbf{0.791} & \textbf{0.191} &  \textbf{4.19} 
                          &           2.65M & \textbf{27.89} & \textbf{0.878} & \textbf{0.101} &  \textbf{4.77} 
                          &           3.03M & \textbf{27.42} & \textbf{0.812} & \textbf{0.182} &  \textbf{3.25} \\
        \midrule
        Taming-3DGS~\cite{mallick2024taming3dgshighqualityradiance}
                          &  \textbf{3.85M} &          25.58 &          0.766 &          0.226 &           6.95 
                          &           3.07M &          27.63 & \textbf{0.869} & \textbf{0.110} &           6.89 
                          &           3.73M &          26.66 &          0.775 &          0.220 &           5.26 \\
        +Ours
                          &           3.87M & \textbf{25.81} & \textbf{0.782} & \textbf{0.210} &  \textbf{4.38} 
                          &  \textbf{2.46M} & \textbf{27.91} &          0.868 &          0.119 &  \textbf{3.80} 
                          &  \textbf{3.14M} & \textbf{27.34} & \textbf{0.792} & \textbf{0.211} &  \textbf{2.71} \\
        \midrule
        \midrule
        \multirow{2}{*}{Method} 
        & \multicolumn{5}{c}{flowers} & \multicolumn{5}{c}{treehill} & \multicolumn{5}{c}{room} \\
        \cmidrule(l{2pt}r{2pt}){2-6} \cmidrule(l{2pt}r{2pt}){7-11} \cmidrule(l{2pt}r{2pt}){12-16}
        & $\mathrm{N_{GS}}\downarrow$ & PSNR$\uparrow$ & SSIM$\uparrow$ & LPIPS$\downarrow$ & Time$\downarrow$ 
        & $\mathrm{N_{GS}}\downarrow$ & PSNR$\uparrow$ & SSIM$\uparrow$ & LPIPS$\downarrow$ & Time$\downarrow$ 
        & $\mathrm{N_{GS}}\downarrow$ & PSNR$\uparrow$ & SSIM$\uparrow$ & LPIPS$\downarrow$ & Time$\downarrow$ \\
        \midrule
        3DGS~\cite{kerbl3Dgaussians}
                          &           3.36M &          21.86 &          0.622 &          0.328 &          16.50 
                          &  \textbf{3.62M} &          22.85 &          0.652 &          0.317 &          17.05 
                          &           1.51M & \textbf{31.72} & \textbf{0.927} & \textbf{0.191} &          15.60 \\
        +Ours
                          &  \textbf{3.22M} & \textbf{22.17} & \textbf{0.631} & \textbf{0.318} & \textbf{10.78} 
                          &           3.63M & \textbf{23.14} & \textbf{0.659} & \textbf{0.309} & \textbf{12.13} 
                          &  \textbf{1.29M} &          31.58 &          0.923 &          0.203 & \textbf{8.06} \\
        \midrule
        Mip-Splatting~\cite{Yu2024MipSplatting}
                          &           4.32M &          22.06 & \textbf{0.656} & \textbf{0.266} &          23.80 
                          &           5.01M &          22.61 &          0.655 & \textbf{0.269} &          26.12 
                          &           2.10M &          31.94 & \textbf{0.933} & \textbf{0.175} &          21.07 \\
        +Ours
                          &  \textbf{3.88M} & \textbf{22.13} &          0.648 &          0.282 & \textbf{13.76} 
                          &  \textbf{4.60M} & \textbf{23.15} & \textbf{0.662} &          0.285 & \textbf{15.47} 
                          &  \textbf{1.59M} & \textbf{32.08} &          0.931 &          0.183 &  \textbf{9.90} \\
        \midrule
        Revising-3DGS*~\cite{bulò2024revisingdensificationgaussiansplatting}
                          &           2.05M &          21.56 &          0.625 &          0.286 &           5.46 
                          &           2.17M &          22.65 &          0.647 &          0.307 &           4.80 
                          &           1.00M &          30.81 &          0.922 & \textbf{0.192} &           4.92 \\
        +Ours
                          &           2.05M & \textbf{22.05} & \textbf{0.648} & \textbf{0.279} &  \textbf{3.88} 
                          &           2.17M & \textbf{22.80} & \textbf{0.656} & \textbf{0.300} &  \textbf{3.30} 
                          &           1.00M & \textbf{31.97} & \textbf{0.929} &          0.194 &  \textbf{2.53} \\
        \midrule
        Taming-3DGS~\cite{mallick2024taming3dgshighqualityradiance}
                          &           2.38M &          21.83 &          0.615 &          0.336 &           4.72 
                          &  \textbf{2.52M} &          22.92 &          0.649 &          0.330 &           4.81 
                          &           1.11M &          31.51 & \textbf{0.925} & \textbf{0.198} &           4.32 \\
        +Ours
                          &  \textbf{2.34M} & \textbf{22.31} & \textbf{0.633} & \textbf{0.320} &  \textbf{3.30} 
                          &           2.88M & \textbf{23.33} & \textbf{0.658} & \textbf{0.315} &  \textbf{3.60} 
                          &   \textbf{0.90M} & \textbf{31.85} &          0.922 &          0.205 &  \textbf{2.38} \\
        \midrule
        \midrule
        \multirow{2}{*}{Method} 
        & \multicolumn{5}{c}{kitchen} & \multicolumn{5}{c}{counter} & \multicolumn{5}{c}{bonsai} \\
        \cmidrule(l{2pt}r{2pt}){2-6} \cmidrule(l{2pt}r{2pt}){7-11} \cmidrule(l{2pt}r{2pt}){12-16}
        & $\mathrm{N_{GS}}\downarrow$ & PSNR$\uparrow$ & SSIM$\uparrow$ & LPIPS$\downarrow$ & Time$\downarrow$ 
        & $\mathrm{N_{GS}}\downarrow$ & PSNR$\uparrow$ & SSIM$\uparrow$ & LPIPS$\downarrow$ & Time$\downarrow$ 
        & $\mathrm{N_{GS}}\downarrow$ & PSNR$\uparrow$ & SSIM$\uparrow$ & LPIPS$\downarrow$ & Time$\downarrow$ \\
        \midrule
        3DGS~\cite{kerbl3Dgaussians}
                          &           1.74M &          31.41 & \textbf{0.933} & \textbf{0.113} &          18.28 
                          &           1.17M & \textbf{29.12} & \textbf{0.915} & \textbf{0.178} &          15.08 
                          &           1.24M & \textbf{32.19} & \textbf{0.947} & \textbf{0.173} &          13.56 \\
        +Ours
                          &  \textbf{1.39M} & \textbf{31.58} &          0.926 &          0.128 &  \textbf{9.68} 
                          &  \textbf{1.01M} &          28.90 &          0.910 &          0.188 &  \textbf{7.31} 
                          &  \textbf{1.06M} &          32.01 &          0.944 &          0.177 &  \textbf{7.28} \\
        \midrule
        Mip-Splatting~\cite{Yu2024MipSplatting}
                          &           2.13M & \textbf{31.86} & \textbf{0.936} & \textbf{0.106} &          23.57 
                          &           1.47M & \textbf{29.33} & \textbf{0.920} & \textbf{0.165} &          19.72 
                          &           1.61M & \textbf{32.41} & \textbf{0.951} & \textbf{0.157} &          18.01 \\
        +Ours
                          &  \textbf{1.71M} &          31.76 &          0.930 &          0.119 & \textbf{12.01} 
                          &  \textbf{1.28M} &          29.22 &          0.915 &          0.176 &  \textbf{9.14} 
                          &  \textbf{1.35M} &          32.32 &          0.946 &          0.164 &  \textbf{9.21} \\
        \midrule
        Revising-3DGS*~\cite{bulò2024revisingdensificationgaussiansplatting}
                          &           1.43M & \textbf{31.93} & \textbf{0.934} & \textbf{0.114} &           7.08 
                          &           0.89M & \textbf{29.23} & \textbf{0.916} & \textbf{0.175} &           6.51 
                          &           1.01M & \textbf{32.26} & \textbf{0.947} & \textbf{0.166} &           5.11 \\
        +Ours
                          &           1.43M &          31.78 &          0.930 &          0.123 &  \textbf{3.66} 
                          &           0.89M &          29.03 &          0.912 &          0.185 &  \textbf{3.01} 
                          &           1.01M &          32.03 &          0.945 &          0.169 &  \textbf{2.58} \\
        \midrule
        Taming-3DGS~\cite{mallick2024taming3dgshighqualityradiance}
                          &           1.68M &          31.09 & \textbf{0.930} & \textbf{0.117} &           7.71 
                          &           1.02M &          29.10 & \textbf{0.914} & \textbf{0.180} &           4.92 
                          &           1.16M & \textbf{32.20} &          0.943 & \textbf{0.175} &           4.00 \\
        +Ours
                          &  \textbf{1.21M} & \textbf{31.52} &          0.926 &          0.127 &  \textbf{3.89} 
                          &  \textbf{0.78M} & \textbf{29.10} &          0.910 &          0.188 &  \textbf{2.63} 
                          &  \textbf{0.80M} &          32.10 & \textbf{0.943} &          0.175 &  \textbf{2.38} \\
        \bottomrule
    \end{tabular}
    \label{tab:scenewise-m360}
\end{table*}

\section{Implementation Details}
\label{sup:details}
In this section, we elaborate on the implementation of our ground truth downsampling strategy, the modification to the learning rate, the modulation on the resolution scheduler, and how we equip \OURS{} to the each backbones. 

\subsection{Ground Truth Downsampling}
\label{sup:details:gtds}

In the LR optimization stage, \OURS{} downsamples the ground truth to supervise the LR renderings from 3D Gaussians. 
We choose the official anti-alias downsampling algorithm from Pytorch to downsample the ground truth images, preventing possible misguidance from aliased supervision hindering the optimization. 

\subsection{Learning Rate}
\label{sup:details:lr}
Because the optimization scheme in \OURS{} is specifically tailored, the learning rate for 3DGS optimization should be properly scheduled to allow \OURS{} to perform the best. 
For all the experiments, we keep all the hyper-parameters unchanged as in the original 3DGS~\cite{kerbl3Dgaussians} expect the positional learning rate. 
We hold it constant as the initial value in the LR optimization stage, and only allow it to decay from the iteration $i=\argmin_i\left\{\lfloor r^{(i)} \rfloor=1\right\}$.

\subsection{Modulation for Resolution Scheduler}
\label{sup:details:modu}

As we pointed out in the paper, \OURS{} allocates the iterations spent on the LR and HR optimization stage, in order to accelerate the optimization process while maintaining the rendering quality of 3DGS. 
To this end, we further modulate $s_r$ to suppress the LR stage while expanding the HR stage. 
Specifically, we apply the natural logarithm to modulate the relative multiple for different resolutions, resulting in a fraction function 
\begin{equation}
\label{eq:6}
    f(\mathbf{F}, \mathbf{F}_{r_m}, \mathbf{F}_{r_i}) = 
        \frac{\ln\left(\mathcal{X}(\mathbf{F})/\mathcal{X}(\mathbf{F}_{r_i})\right)}
             {\ln\left(\mathcal{X}(\mathbf{F})/\mathcal{X}(\mathbf{F}_{r_m})\right)}, 
\end{equation}
with which we have 
\begin{equation}
\label{eq:7}
\begin{split}
    s_{r_i} &= S\cdot(1 - f(\mathbf{F}, \mathbf{F}_{r_m}, \mathbf{F}_{r_i}))  \\
            &=
    S \cdot 
        \frac{\ln{\left(\mathcal{X}(\mathbf{F}_{r_i})/\mathcal{X}(\mathbf{F}_{r_m})\right)}}
             {\ln{\left(\mathcal{X}(\mathbf{F})/\mathcal{X}(\mathbf{F}_{r_m})\right)}}. 
\end{split}
\end{equation}
\cref{eq:7} is the actual scheduling function we use for \OURS{} and all experiments in the paper. 

\subsection{Densification of DashGaussian}
\label{sup:details:densification}
We explain how a densification step is performed in \OURS{}. 
Given an optimization step $i$ where a densification operation is performed, we set $N_\mathrm{GS}$ as the current number of Gaussian primitives, and $P_i$ as the desired primitive number after this densification operation. 
We first perform a prune operation on the primitives, resulting in $N_\mathrm{GS}'$ primitives left. 
Then we perform clone and split operation to those Gaussian primitives that satisfy the densification condition of the backbone while having a top $P_i-N_\mathrm{GS}'$ densification score. 
Each primitive is cloned or split depending on its size, which follows the same routine as in 3DGS~\cite{kerbl3Dgaussians}. 
When the densification score differs, slight difference is there to adopt DashGaussian to the backbones. 

\subsection{Equipping \OURS{} to Backbones}
\label{sup:details:backbones}

When we enhance a 3DGS backbone with \OURS{}, only few modifications to the codebase are adopted, including a new learning rate schedule (see \cref{sup:details:lr} in supplementary), the rendering resolution schedule, the primitive growth schedule, and a distinct densification operation (see \cref{sup:details:densification}). 
These four modifications from \OURS{} are basically common for different backbones, with slight differences in the densification operation. 
We introduce these differences in detail below. 

\paragraph{3DGS~\cite{kerbl3Dgaussians}}
\vspace{-4mm}
The original 3DGS has a densification score defined as the positional gradient of each primitive. 
We keep the densification score, and follow \cref{sup:details:densification} to perform densification. 
The optimizer is Adam~\cite{KingBa15}.

\paragraph{Taming-3DGS~\cite{mallick2024taming3dgshighqualityradiance}}
\vspace{-4mm}
At the moment of this paper is written, the codebase of Taming-3DGS has an efficient backward pass implementation with the rest same as 3DGS~\cite{kerbl3Dgaussians}. 
So the densification operation is the same as \cref{sup:details:densification}.
The optimizer is Sparse Adam~\cite{mallick2024taming3dgshighqualityradiance}.

\paragraph{Revising-3DGS*~\cite{bulò2024revisingdensificationgaussiansplatting}}
\vspace{-4mm}
Because Revising-3DGS is not open-sourced, we reproduce it on top of the codebase of Taming-3DGS~\cite{mallick2024taming3dgshighqualityradiance}. 
Notice that, Revising-3DGS demands appointing the primitive number in the final 3DGS model. 
To this end, we follow~\cite{bulò2024revisingdensificationgaussiansplatting} to set $P_\mathrm{fin}$ as a constant, with our momentum-based primitive budgeting disabled. 
Revising-3DGS has its densification score defined as rendering error, thus the densification operation follows \cref{sup:details:densification} the same. 
The optimizer is Sparse Adam~\cite{mallick2024taming3dgshighqualityradiance}.

\paragraph{Mip-Splatting~\cite{Yu2024MipSplatting}}
\vspace{-4mm}
Mip-Splatting defines its densification scores as two different scalars, where these two scores are used together to specify the primitives to be densified. 
For the primitive selection in densification, we merge the primitives selected with these two densification scores as a whole set and randomly select $P_i-N_\mathrm{GS}'$ primitives from them for densification. 
The optimizer is Adam~\cite{KingBa15}.

\section{Primitive Growth of \OURS{}}
\label{sup:primgrowth}

\subsection{Why Fewer Primitives in DashGaussian?}
\label{sup:primgrowth:fewer}
As we discussed in \cref{sup:details:densification} of the supplementary, the densified primitives are selected by the vanilla densification score together with a top-$k$ selection to control the growth of primitives. 
In the early LR optimization stage, the top-$k$ selection prevents the situation where a large amount of Gaussians pass the threshold resulting in an explosion of primitive growth.
In the late HR optimization stage, the optimized 3DGS is relatively stable and there are less than $P_i-N_\mathrm{GS}'$ primitives passing over the densification threshold. 
The threshold-based densification and the top-$k$ selection constrain each other, resulting in less densified primitives in the final optimized 3DGS than $P_\mathrm{fin}$. 

\subsection{Constant Primitive Number}
\label{sup:primgrowth:constant}
Intuitively, $P_\mathrm{fin}$, as the primitive number upper-bound, is likely to influence the size of the optimized 3DGS. 
When $P_\mathrm{fin}$ is extremely large, the primitive number $N_\mathrm{GS}$ in the optimized 3DGS is likely to explode, resulting in out-of-memory (OOM) issue. 
Nonetheless, thanks to the double-condition densification scheme as discussed in \cref{sup:primgrowth:fewer}, we can prevent OOM when $P_\mathrm{fin}$ is arbitrarily large. 
Concretely, as \cref{tab:ablation-gamma} in the supplementary shows, when we remove the primitive scheduler, which is equivalent to setting $P_\mathrm{fin}$ as infinity, $N_\mathrm{GS}$ still approximates to the primitive number of the backbone. 
This indicates the primitive growth of \OURS{} is stable and safe. 
Also, one can simply remove the threshold condition and only keep the top-k selection in densification, such that $N_\mathrm{GS}$ will exactly equal to $P_\mathrm{fin}$ after the optimization.

\begin{table*}[t]
    \centering
    \footnotesize
    \setlength\tabcolsep{2pt}
    \captionsetup{singlelinecheck=false}
    \caption{
        Scene-wise quantitative results over Deep Blending dataset~\cite{DeepBlending2018}. 
    }
    \begin{tabular} {l | rrrrr | rrrrr}
        \toprule
        \multirow{2}{*}{Method} 
        & \multicolumn{5}{c}{drjohnson} & \multicolumn{5}{c}{playroom}\\
        \cmidrule(l{2pt}r{2pt}){2-6} \cmidrule(l{2pt}r{2pt}){7-11}
        & $\mathrm{N_{GS}}\downarrow$ & PSNR$\uparrow$ & SSIM$\uparrow$ & LPIPS$\downarrow$ & Time$\downarrow$ 
        & $\mathrm{N_{GS}}\downarrow$ & PSNR$\uparrow$ & SSIM$\uparrow$ & LPIPS$\downarrow$ & Time$\downarrow$ \\
        \midrule
        3DGS~\cite{kerbl3Dgaussians}
                          &           3.31M & \textbf{29.09} & \textbf{0.901} & \textbf{0.244} &          19.47 
                          &           2.32M & \textbf{29.90} & \textbf{0.907} & \textbf{0.244} &          17.27 \\
        +Ours
                          &  \textbf{2.47M} &          28.75 &          0.895 &          0.262 &  \textbf{8.68} 
                          &  \textbf{1.89M} &          29.84 &          0.906 &          0.249 &  \textbf{7.64} \\
        \midrule
        Mip-Splatting~\cite{Yu2024MipSplatting}
                          &           4.13M &          28.66 &          0.898 & \textbf{0.243} &          26.80 
                          &           2.83M &          29.85 &          0.906 & \textbf{0.235} &          20.33 \\
        +Ours
                          &  \textbf{3.04M} & \textbf{28.80} &          0.899 &          0.255 & \textbf{10.73} 
                          &  \textbf{2.06M} & \textbf{30.29} & \textbf{0.906} &          0.241 &  \textbf{8.86} \\
        \midrule
        Revising-3DGS*~\cite{bulò2024revisingdensificationgaussiansplatting}
                          &           2.52M &          29.12 &          0.902 & \textbf{0.248} &           4.69 
                          &           1.51M &          29.96 &          0.909 & \textbf{0.248} &           3.43 \\
        +Ours
                          &           2.52M & \textbf{29.35} & \textbf{0.903} &          0.252 &  \textbf{2.47} 
                          &           1.51M & \textbf{30.18} & \textbf{0.911} &          0.250 &  \textbf{1.94} \\
        \midrule
        Taming-3DGS~\cite{mallick2024taming3dgshighqualityradiance}
                          &           2.93M &          29.24 &          0.905 & \textbf{0.241} &           5.38 
                          &           1.72M &          30.14 & \textbf{0.908} &          0.249 &           3.65 \\
        +Ours
                          &  \textbf{2.53M} & \textbf{29.56} & \textbf{0.905} &          0.247 &  \textbf{2.49} 
                          &  \textbf{1.36M} & \textbf{30.49} &          0.908 & \textbf{0.249} &  \textbf{1.91} \\
        \bottomrule
    \end{tabular}
    \label{tab:scenewise-db}
\end{table*}
\begin{table*}[t]
    \centering
    \footnotesize
    \setlength\tabcolsep{2pt}
    \captionsetup{singlelinecheck=false}
    \caption{
        Scene-wise quantitative results over Tanks\&Temples dataset~\cite{Knapitsch2017}. 
    }
    \begin{tabular} {l | rrrrr | rrrrr}
        \toprule
        \multirow{2}{*}{Method} 
        & \multicolumn{5}{c}{train} & \multicolumn{5}{c}{truck}\\
        \cmidrule(l{2pt}r{2pt}){2-6} \cmidrule(l{2pt}r{2pt}){7-11}
        & $\mathrm{N_{GS}}\downarrow$ & PSNR$\uparrow$ & SSIM$\uparrow$ & LPIPS$\downarrow$ & Time$\downarrow$ 
        & $\mathrm{N_{GS}}\downarrow$ & PSNR$\uparrow$ & SSIM$\uparrow$ & LPIPS$\downarrow$ & Time$\downarrow$ \\
        \midrule
        3DGS~\cite{kerbl3Dgaussians}
                          &           1.08M &          21.78 & \textbf{0.812} & \textbf{0.207} &           8.46 
                          &           2.58M &          25.45 & \textbf{0.882} & \textbf{0.146} &          12.72 \\
        +Ours
                          &  \textbf{1.01M} & \textbf{22.01} &          0.806 &          0.223 &  \textbf{5.91} 
                          &  \textbf{1.85M} & \textbf{25.66} &          0.882 &          0.156 &  \textbf{6.38} \\
        \midrule
        Mip-Splatting~\cite{Yu2024MipSplatting}
                          &           1.47M &          21.76 & \textbf{0.825} & \textbf{0.190} &          11.31 
                          &           3.24M &          25.73 & \textbf{0.892} & \textbf{0.123} &          17.15 \\
        +Ours
                          &  \textbf{1.18M} & \textbf{22.01} &          0.820 &          0.207 &  \textbf{7.05} 
                          &  \textbf{2.07M} & \textbf{25.91} &          0.892 &          0.137 &  \textbf{8.07} \\
        \midrule
        Revising-3DGS*~\cite{bulò2024revisingdensificationgaussiansplatting}
                          &           0.96M &          22.10 & \textbf{0.823} & \textbf{0.211} &           4.18 
                          &           1.68M & \textbf{25.65} & \textbf{0.889} & \textbf{0.126} &           4.81 \\
        +Ours
                          &           0.96M & \textbf{22.13} &          0.817 &          0.223 &  \textbf{2.88} 
                          &           1.68M &          25.56 &          0.888 &          0.129 &  \textbf{3.43} \\
        \midrule
        Taming-3DGS~\cite{mallick2024taming3dgshighqualityradiance}
                          &           1.07M &          21.96 &          0.816 & \textbf{0.204} &           3.64 
                          &           1.96M &          25.28 &          0.882 & \textbf{0.145} &           4.40 \\
        +Ours
                          &  \textbf{1.00M} & \textbf{22.22} & \textbf{0.817} &          0.208 &  \textbf{2.75} 
                          &  \textbf{1.39M} & \textbf{25.71} & \textbf{0.884} &          0.153 &  \textbf{2.48} \\
        \bottomrule
    \end{tabular}
    \label{tab:scenewise-tat}
\end{table*}

\begin{table}[t!]
    \footnotesize
    \centering
    \captionsetup{singlelinecheck=false}
    \caption{
        Quantitative results of the ablation study over $\gamma$ on Mip-NeRF 360 dataset~\cite{Barron_2022_CVPR}, with Taming-3DGS~\cite{mallick2024taming3dgshighqualityradiance} as the backbone of DashGaussian. 
        `$1^-$' indicates where $\gamma$ infinitely approaches $1$ from below, equivalent to removing the primitive scheduler. 
    }
    \setlength{\tabcolsep}{3pt}
    \begin{tabular} {l | rrrrr}
        \toprule
        $\gamma$                  & $N_\mathrm{GS}\downarrow$ & PSNR$\uparrow$ & SSIM$\uparrow$ & LPIPS$\downarrow$ & Time$\downarrow$ \\
        \midrule
        Taming-3DGS~\cite{mallick2024taming3dgshighqualityradiance}
                                  &             2.02M &          27.61 &          0.821 &             0.210 &             5.51 \\
        \midrule
        $1^-$                     &             2.46M & \textbf{27.99} & \textbf{0.829} &    \textbf{0.196} &             4.43 \\
        $0.99$                    &             2.17M &          27.98 &          0.828 &             0.204 &             3.49 \\
        $0.98$ (Ours)             &             2.04M &          27.92 &          0.826 &             0.208 &             3.23 \\
        $0.95$                    &             1.52M &          27.81 &          0.820 &             0.218 &             2.77 \\
        $0.90$                    &    \textbf{1.09M} &          27.67 &          0.809 &             0.237 &    \textbf{2.41} \\
        \bottomrule
    \end{tabular}
    \label{tab:ablation-a}
\end{table}
\begin{table}[t]
    \footnotesize
    \centering
    \captionsetup{singlelinecheck=false}
    \caption{
        Quantitative results of the ablation study over $a$ on Mip-NeRF 360 dataset~\cite{Barron_2022_CVPR}, with Taming-3DGS~\cite{mallick2024taming3dgshighqualityradiance} as the backbone of DashGaussian. 
        $a=1$ equals removing the resolution scheduler, which is the same as `+Prim-Sche.' in Tab.~3 of the paper. 
    }
    \setlength{\tabcolsep}{3pt}
    \begin{tabular} {l | rrrrr}
        \toprule
        $a$                  & $N_\mathrm{GS}\downarrow$ & PSNR$\uparrow$ & SSIM$\uparrow$ & LPIPS$\downarrow$ & Time$\downarrow$ \\
        \midrule
        Taming-3DGS~\cite{mallick2024taming3dgshighqualityradiance}
                               &             2.02M &          27.61 &          0.821 &             0.210 &             5.51 \\
        \midrule
        $1$                    &             2.23M &          27.85 &          0.824 &             0.208 &             4.60 \\
        $2$                    &             2.05M & \textbf{27.97} & \textbf{0.827} &    \textbf{0.204} &             4.62 \\
        $4$ (Ours)             &             2.04M &          27.92 &          0.826 &             0.208 &             3.23 \\
        $6$                    &             2.00M &          27.93 &          0.825 &             0.211 &             2.88 \\
        $10$                   &    \textbf{1.95M} &          27.91 &          0.824 &             0.213 &    \textbf{2.79} \\
        \bottomrule
    \end{tabular}
    \label{tab:ablation-gamma}
\end{table}

\section{Scene-wise Qualitative Results}
\label{sup:scene-wise}
As complementary to the Tab.~2 in the paper, we report scene-wise quantitative results on all three datasets in \cref{tab:scenewise-m360}, \cref{tab:scenewise-db} and \cref{tab:scenewise-tat}, respectively.

\section{Ablation on Hyper-parameters}
\label{sup:ablation-param}
In this section, we discuss how the hyper-parameters influence the performance of \OURS{}. 

\paragraph{Momentum-based Primitive Budgeting.}
\vspace{-3mm}
The two hyper-parameters in Eq.~5 of the paper, $\gamma$ and $\eta$, directly affect $P_\mathrm{fin}$ and thus $N_\mathrm{GS}$. 
Before the ablation experiment, we first analyze the inner connection between $\gamma, \eta$, and $P_\mathrm{fin}$ to show how we select them. 

Consider when Eq.~5 of the paper is stable and $P_\mathrm{fin}$ converges to a constant, we can rewrite Eq.~5 of the paper, resulting in
\begin{equation}
    \label{eq:8}
    P_\mathrm{fin} = \frac{\eta}{1-\gamma}P_\mathrm{add}^{(i)}, 
\end{equation}
which reveals the inner connection between $P_\mathrm{fin}, \gamma, \eta$ and $P_\mathrm{add}^{(i)}$. 
With the above equation, one can select $\gamma$ and $\eta$ based on the typical $P_\mathrm{add}^{(i)}$ to expect a maximum $P_\mathrm{fin}$. 
However, as introduced in \cref{sup:primgrowth:constant} of the supplementary, $N_\mathrm{GS}$ is stable with arbitrarily large $P_\mathrm{fin}$. 
So selecting $\gamma$ and $\eta$ is to make $P_\mathrm{fin}$ approximate to $N_\mathrm{GS}$, which effectively functions the primitive scheduler to accelerate the optimization while preventing under reconstruction with an over small $P_\mathrm{fin}$.

Since changing $\gamma$ and $\eta$ is essentially equivalent to changing the one parameter $\eta / (1 - \gamma)$ in \cref{eq:8}, we only perform ablation on $\gamma$. 
The results are reported in \cref{tab:ablation-gamma}.
As analyzed in \cref{eq:8}, a larger $\gamma$ indicates a larger $P_\mathrm{fin}$, and thus a bigger $N_\mathrm{GS}$, higher rendering quality, and more optimization time. 
To notice, even when $\gamma=1^-$ where $P_\mathrm{fin}$ is infinite (implemented as removing the primitive scheduler), the value of $N_\mathrm{GS}$ is stable and significant acceleration brought by our method is still observed.

\paragraph{Initial Resolution.}
\vspace{-4mm}
The hyper-parameter $a$ in Sec.~4.2 of the paper decides the initial rendering resolution at the beginning of optimization, where a larger $a$ indicates a smaller initial rendering resolution. 
We report ablation on $a$ in \cref{tab:ablation-a}. 
Results show that the rendering quality is basically insensitive to $a$, while the optimization time significantly reduces with the increase of $a$.

\section{Evaluation on Large-scale Reconstruction}
\label{sup:largescale}

We conduct evaluation for \OURS{} on large-scale reconstruction with the MatrixCity~\cite{Li_2023_ICCV} dataset, where the backbone is Taming-3DGS~\cite{mallick2024taming3dgshighqualityradiance} and the divide-and-conquer strategy follows VastGaussian~\cite{Lin_2024_CVPR} to split the scene into 9 blocks, denoted as LargeBase in \cref{tab:largescale}. 
The average training time on 9 blocks is reported (rendered under 1080P). 
While slightly improving the reconstruction quality, \OURS{} helps the backbone to reduce the training time by 41.6\% on each block. 
This experiment again strongly supports our claim that \OURS{} can serve as a generalized plugin to accelerate the optimization of different 3DGS backbones and can well scale up from room-scale to large-scale reconstruction.

\begin{table}[t]

    \centering
    \captionsetup{singlelinecheck=false}
    \caption{
        The report of large-scale reconstruction results on MatrixCity dataset. 
        Time is reported in minutes. 
    }
    \vspace{-5pt}
    \resizebox{\linewidth}{!}{
        \begin{tabular}{l | rrrrr}
            \toprule
            Method         & $\mathrm{N_{GS}}\downarrow$ & PSNR$\uparrow$ & SSIM$\uparrow$ & LPIPS$\downarrow$ & Time$\downarrow$ \\
            \midrule
            LargeBase      &                    18.28M &          26.51 &          0.858 &             0.186 &           44.22 \\
            +Ours          &           \textbf{18.08M} & \textbf{26.59} & \textbf{0.865} &    \textbf{0.182} &  \textbf{25.82} \\
            \bottomrule
        \end{tabular}
    }
    \vspace{-5pt}
    \label{tab:largescale}
\end{table}


\end{document}